\documentclass{article}
\usepackage[preprint]{log_2025}						

\usepackage{booktabs}						
\usepackage{multirow}						
\usepackage{amsfonts}						
\usepackage{graphicx}						
\usepackage{duckuments}						

\usepackage{makecell}
\usepackage{booktabs}
\usepackage{siunitx}
\usepackage[numbers,compress,sort]{natbib}	

\usepackage[utf8]{inputenc} 
\usepackage[T1]{fontenc}    

\usepackage{url}            
\usepackage{booktabs}       
\usepackage{amsfonts}       
\usepackage{nicefrac}       
\usepackage{microtype}      
\usepackage{xcolor}         

\usepackage{pythonhighlight}
\usepackage{graphicx}
\usepackage{multirow}
\usepackage{multicol}
\usepackage{subcaption}

\usepackage{wrapfig}

\usepackage{diagbox}
\usepackage{pifont}

\definecolor{redx}{RGB}{180,0,0}
\definecolor{greenx}{RGB}{0,180,0}

\newcommand{\redxmark}{\color{redx}\ding{55}}
\newcommand{\greencmark}{\color{greenx}\ding{51}}
\definecolor{redx}{RGB}{180,0,0}
\definecolor{greenx}{RGB}{0,180,0}

\usepackage{tikz}
\definecolor{yellowx}{RGB}{230,180,0}

\newcommand{\yellowcircle}{%
  \tikz[baseline=-0.6ex]%
    \draw[line width=1pt, draw={rgb,255:red,230;green,180;blue,0}, fill=white]
      (0,0) circle (0.8ex);%
}

\usepackage{listings}
\usepackage{xcolor}

\newif\ifshowcomment
\showcommenttrue

\ifshowcomment

\fi

\title[FedGraph: A Research Library and Benchmark for Federated Graph Learning]{FedGraph: A Research Library and Benchmark for Federated Graph Learning}

\author[Yao et al.]{
Yuhang Yao$^{1}$, 
Yuan Li$^{1}$, 
Xinyi Fan$^{1}$, 
Junhao Li$^{1}$, 
Kay Liu$^{2}$, 
Weizhao Jin$^{3}$, 
Yu Yang$^{1}$,\\
\textbf{Srivatsan Ravi$^{3}$, 
Philip S. Yu$^{2}$, 
Carlee Joe-Wong$^{1}$}\\
$^{1}$Carnegie Mellon University \quad
$^{2}$University of Illinois Chicago \quad
$^{3}$University of Southern California
\\
\texttt{yuhangyao8@gmail.com}
}

\begin{document}

\maketitle

\begin{abstract}
Federated graph learning is an emerging field with significant practical challenges. While algorithms have been proposed to improve the accuracy of training graph neural networks, such as node classification on federated graphs, the system performance is often overlooked, despite it is crucial for real-world deployment. To bridge this gap, we introduce FedGraph, a research library designed for practical distributed training and comprehensive benchmarking of FGL algorithms. FedGraph supports a range of state-of-the-art graph learning methods and includes a monitoring class that evaluates system performance, with a particular focus on communication and computation costs during training. Unlike existing federated learning platforms, FedGraph natively integrates homomorphic encryption to enhance privacy preservation and supports scalable deployment across multiple physical machines with system-level performance evaluation to guide the system design of future algorithms. To enhance efficiency and privacy, we propose a low-rank communication scheme for algorithms like FedGCN that require pre-training communication, accelerating both the pre-training and training phases. Extensive experiments benchmark FGL algorithms on three major graph learning tasks and demonstrate FedGraph as the first efficient FGL framework to support encrypted low-rank communication and scale to graphs with 100 million nodes. 

\end{abstract}


\section{Introduction}
\begin{figure}[t]
    \centering
    \includegraphics[width=0.97\textwidth]{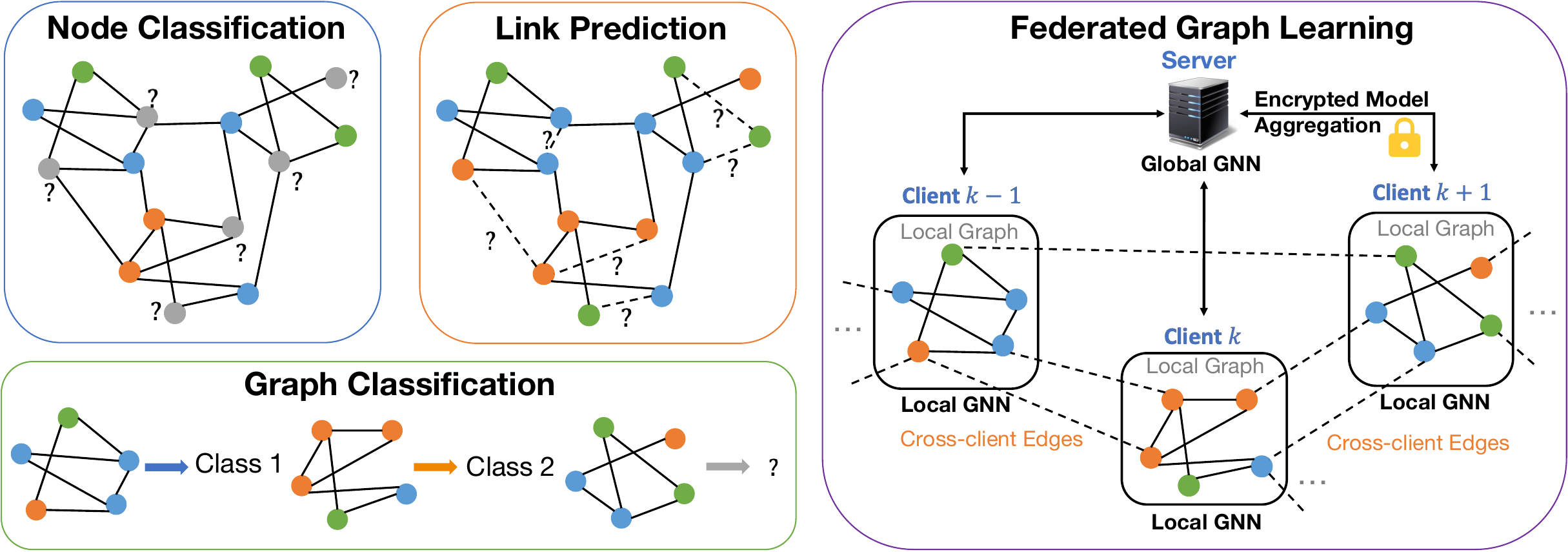}
    \caption{Modeling Applications as Graph Tasks (left) and Federated Graph Training Schematic (right). Node classification aims to predict the labels of grey nodes using structural and feature information from neighboring nodes. Link prediction involves inferring potential future edges between nodes based on existing graph structure and node features. Graph classification assigns labels to entire graphs based on their topology and node information. In federated graph learning, nodes are located across multiple clients with potential cross-client edges. Each client trains a local GNN on its subgraph and exchanges encrypted model updates with a central server, enabling collaborative learning without sharing raw data.} 
    \label{fig:main_fedgraph_problem}
    \vspace{-4mm}
\end{figure}

Graph neural networks aim to learn representations of graph-structured data that capture features associated with graph nodes as well as edges between them~\citep{bronstein2017geometric}. 
Most graph applications can modeled as one of three major graph learning problems: node classification (e.g., classifying nodes representing papers in citation networks based on their research topic), link prediction (e.g., recommending the formation of links that represent friendship between users in social networks), or graph classification (e.g. classifying types of proteins in biology, where each protein is represented as a graph). Figure~\ref{fig:main_fedgraph_problem} (left) illustrates these three types of graph learning tasks~\citep{benamira2019semi,zhang2020semi}.

In practice, graph data is often too large to be trained on a single server or may naturally exist on multiple local clients. For example, graph learning on records of billions of users' website visits requires significant computational resources, beyond those of a single server. Even if a single server or data center could hold such information, privacy regulations may require that it be stored where it was generated, e.g., the General Data Protection Regulation (GDPR) in Europe and Payment Aggregators and Payment Gateways (PAPG) in India prevent private user data from being shared across international borders. 
Users may also not want to share their personal data with an external server. In response to these concerns, federated learning has been widely studied as a way to preserve user privacy while training accurate models on data stored at multiple clients~\citep{zhao2018federated}.

These challenges to training on real-world graphs motivate \textit{Federated Graph Learning} (FGL) as an important research topic~\citep{liu2024federated}. The training schematic of such a federated architecture is shown in Figure~\ref{fig:main_fedgraph_problem} (right): each client maintains a local graph and a local Graph Neural Network (GNN) model, which can be aggregated with other clients' local models at a coordinator server. Within this architecture, there are many different FGL algorithms that handle local model updates and aggregation differently, resulting in varying model accuracies and system performance (e.g., runtime, communication costs)~\citep{xie2021federated,zhang2021subgraph,yao2024fedgcn}. 
While evaluating the model accuracy of such algorithms can be easily done with simulations on open graph datasets~\citep{hu2020open}, evaluating their \textit{real-world system performance} requires sophisticated benchmarking platforms that can fairly compare multiple FGL algorithms.

Vanilla FGL benchmark platforms are designed for research purposes only and simulate the actions of multiple clients on a single machine~\citep{xie2021federated,li2024openfgl}. Thus, they do not capture the real communication and computation costs of training graph models in a federated manner across distinct physical clients. Meanwhile, existing libraries mainly focus on overcoming client data heterogeneity, with limited support for GNN training. For example, FedScale~\citep{lai2022fedscale} does not support graph models, while FedGraphNN~\citep{he2021fedgraphnn} and FederatedScope-GNN~\citep{wang2022federatedscope} are not well-maintained and only support basic federated learning methods like FedAvg. Users need to implement the FGL algorithms themselves, including steps unique to FGL tasks such as handling cross-client graph edges that begin and end at different clients. Moreover, none of these platforms natively support FGL enhancements that may be needed in practice, e.g., homomorphic encryption for enhanced privacy~\citep{jin2023fedml,yao2024fedgcn}.  

To meet these shortcomings, we introduce \textit{FedGraph}, a research library to easily train GNNs in federated settings. As shown in Table~\ref{tab:former_work}, FedGraph supports various federated training methods of graph neural networks under both simulated and real federated environments, as well as encrypted communication between clients and the central server for model update and information aggregation.
%
%
%
We summarize the \textbf{contributions} of FedGraph as follows. 
\begin{itemize}
    \item FedGraph is the \textit{first} Python library built for real-world federated graph learning systems, including system optimizations that improve efficiency, scalability, and privacy preservation, as well as multiple state-of-the-art FGL algorithms for easy comparison between them.
    \item FedGraph natively supports \textit{homomorphic encrypted aggregation} to strengthen privacy protection and provides an \textit{advanced system-level monitor class} that enables detailed analysis of communication and computation costs throughout training.
    \item FedGraph proposes a low-rank communication scheme for algorithms like FedGCN to accelerate both pre-train communication and training phases. 
    \item Extensive experiments benchmark system performance on three major FGL tasks and showcase support for a privacy-preserving federated system on graphs with 100 million nodes.
\end{itemize}

\begin{table*}[t]
\caption{Comparison with Existing Frameworks. FedGraph supports distributed FGL, cross-client edges, encrypted aggregation, and system-level profiling for large-scale optimization.}

\centering
\resizebox{\textwidth}{!}
{
\begin{tabular}{|c|c|c|c|c|c|}
\hline
                     & Vanilla FGL & FedScale & FedGraphNN & FederatedScope-GNN  & \textbf{FedGraph}                       \\ \hline
Distributed Training & \redxmark       & \greencmark                  & \greencmark                         & \greencmark                     & \greencmark \\ \hline
 Graph Learning           & \greencmark     & \redxmark                      & \greencmark                       & \greencmark                     & \greencmark \\ \hline
Multiple FGL Algorithms    & \greencmark       & \redxmark                   & \redxmark                         & \yellowcircle                   & \greencmark \\ \hline
Cross-Client Edges    & \redxmark       & \redxmark                   & \redxmark                         & \redxmark                   & \greencmark \\ \hline
Encrypted Aggregation    & \redxmark       & \redxmark                   & \redxmark                         & \redxmark                   & \greencmark \\ \hline
System Level Profiler    & \redxmark       & \redxmark                   & \redxmark                         & \redxmark                   & \greencmark \\ \hline
Large Scale ML Optimizations    & \redxmark       & \redxmark                   & \redxmark                         & \redxmark                   & \greencmark \\ \hline
\end{tabular}
}
\label{tab:former_work}
\vspace{-4mm}
\end{table*}

In this paper, we first overview the system design in Section~\ref{sec:design}, followed by highlighting the key system components in Section~\ref{sec:system}. In Section~\ref{sec:case}, we present a case study demonstrating how FedGraph facilitates the design and test of low-rank pre-training communication in FGL. We then benchmark the performance on three tasks and evaluate its scalability in Sections~\ref{sec:benchmark-tasks}, and conclude in Section~\ref{sec:conclusion}.


\section{FedGraph System Design}\label{sec:design}
In this section, we outline the design principles of the FedGraph library and demonstrate how these principles are implemented in our system design (Figure~\ref{fig:fedgraph_system_design_quickstart}).

\subsection{Design Principles}
\begin{figure}[ht]
    \centering  
    \includegraphics[width = 0.98\textwidth]{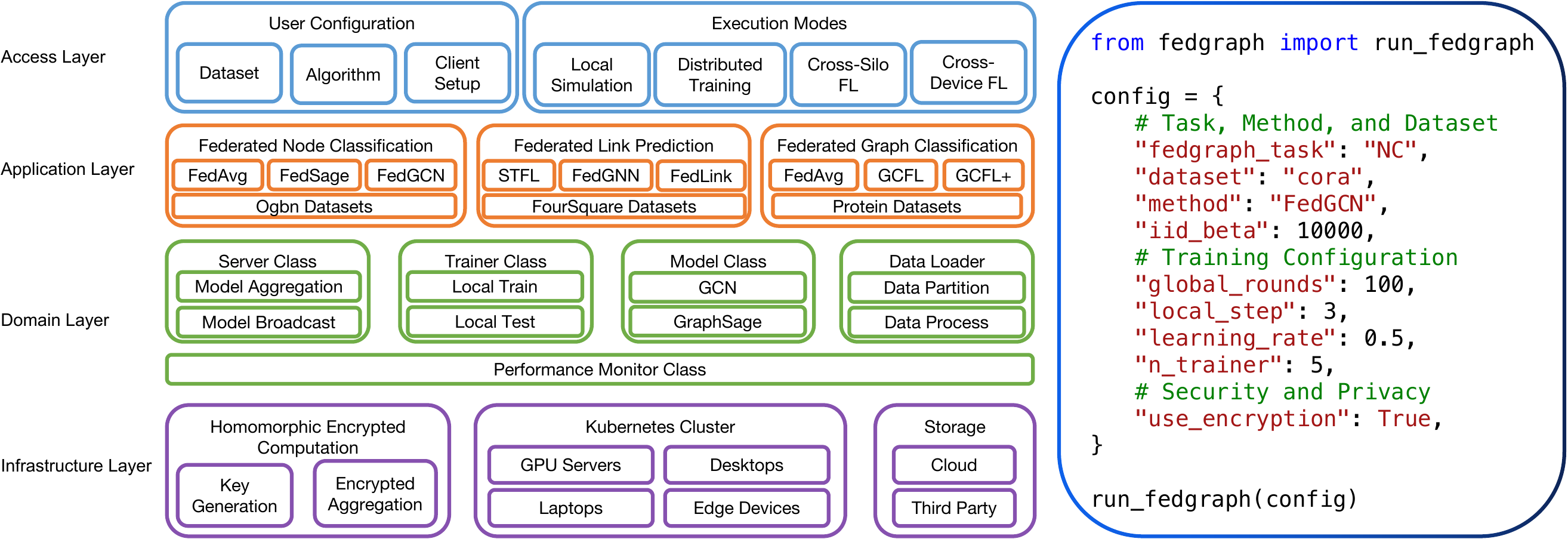}
    \caption{Design Diagram of FedGraph (left) and Quick Start Example (right). The system is organized into four layers: the user access layer, the application layer, the domain layer, and the infrastructure layer. Users only need to focus on the access layer, while the developers can focus on one of the remaining layers based on the domain knowledge.}
    \label{fig:fedgraph_system_design_quickstart}
    \vspace{-4mm}
\end{figure}

The main focus of FedGraph is providing a scalable and privacy-preserving federated graph learning system with ease of use for federated learning researchers and applied scientists in industry. As illustrated in Figure~\ref{fig:fedgraph_system_design_quickstart} (left), the system architecture is structured according to four design principles.

\textbf{Optimized usability with simple configurations for simulation and federated training}: At the access layer, users configure training settings and can seamlessly switch between local simulation and federated training using the same codebase. With just 10-20 lines of code, users can train a federated GNN model, as we discuss further in Section~\ref{subsec:example}. FedGraph abstracts away the complexity of federated graph training, offering a unified platform for training and evaluation.

\textbf{Benchmark existing federated graph learning methods}: In the application layer, FedGraph supports three FGL tasks (node classification, link prediction, and graph classification) and a wide range of state-of-the-art model training algorithms for each task, implemented either by the original authors or library developers. Appendix~\ref{sec:appendix-datasets} lists the currently supported algorithms and datasets.

\textbf{Extensibility to new datasets and algorithms}: In the domain layer, FedGraph separates code into modular components, such as data loaders and training classes. Researchers can extend existing models by inheriting base trainer classes and implementing their own algorithms or data pipelines.

\textbf{Scalability and privacy preservation for real-world deployment}: At the infrastructure layer, FedGraph leverages Ray and Kubernetes to enable distributed computation across edge and cloud environments. Homomorphic encryption can be enabled to secure federated aggregation. This architecture ensures that the system can meet the demands of privacy-preserving, large-scale federated graph learning in realistic settings.

\subsection{FedGraph Use Example}\label{subsec:example}
Researchers can easily install the FedGraph library by running \texttt{pip install fedgraph} in Python, configure experiments, and start training federated GNN models. Appendices~\ref{sec:appendix-code},~\ref{sec:appendix-api}, and~\ref{sec:appendix-workflow} provide more details about the code structure and user API. As illustrated in Figure~\ref{fig:fedgraph_system_design_quickstart} (right), users can initiate a federated graph learning experiment through a straightforward configuration process. \textbf{Task, method, and dataset specification:} The user specifies the learning task, algorithm, dataset, and client data distribution. FedGraph enforces explicit task-method combinations to ensure compatibility and reproducibility. \textbf{Training configuration:} Key hyperparameters, such as the number of global rounds, local training steps, learning rate, and number of participating trainers, are defined in the configuration. \textbf{Security and privacy settings:} Users can enable homomorphic encryption to ensure privacy-preserving training. \textbf{Execution:} Once settled configuration, calling the API function \texttt{run\_fedgraph(config)} launches the experiment. FedGraph then automatically handles data loading, client initialization, and distributed training across clients.

\section{FedGraph System Highlights}\label{sec:system}
In this section, we first introduce FedGraph's monitoring system, which enables usable benchmarking of FGL methods. We then introduce two infrastructure features, FedGraph Homomorphic Encryption for privacy-preserving aggregation and FedGraph Kubernetes for scalable distributed training. Finally, we discuss supported configurations for users to optimize large-scale model training.

\subsection{FedGraph Monitoring System}


\begin{wrapfigure}{r}{0.6\textwidth}
    \centering
    \vspace{-10mm}
    \includegraphics[width=0.6\textwidth]{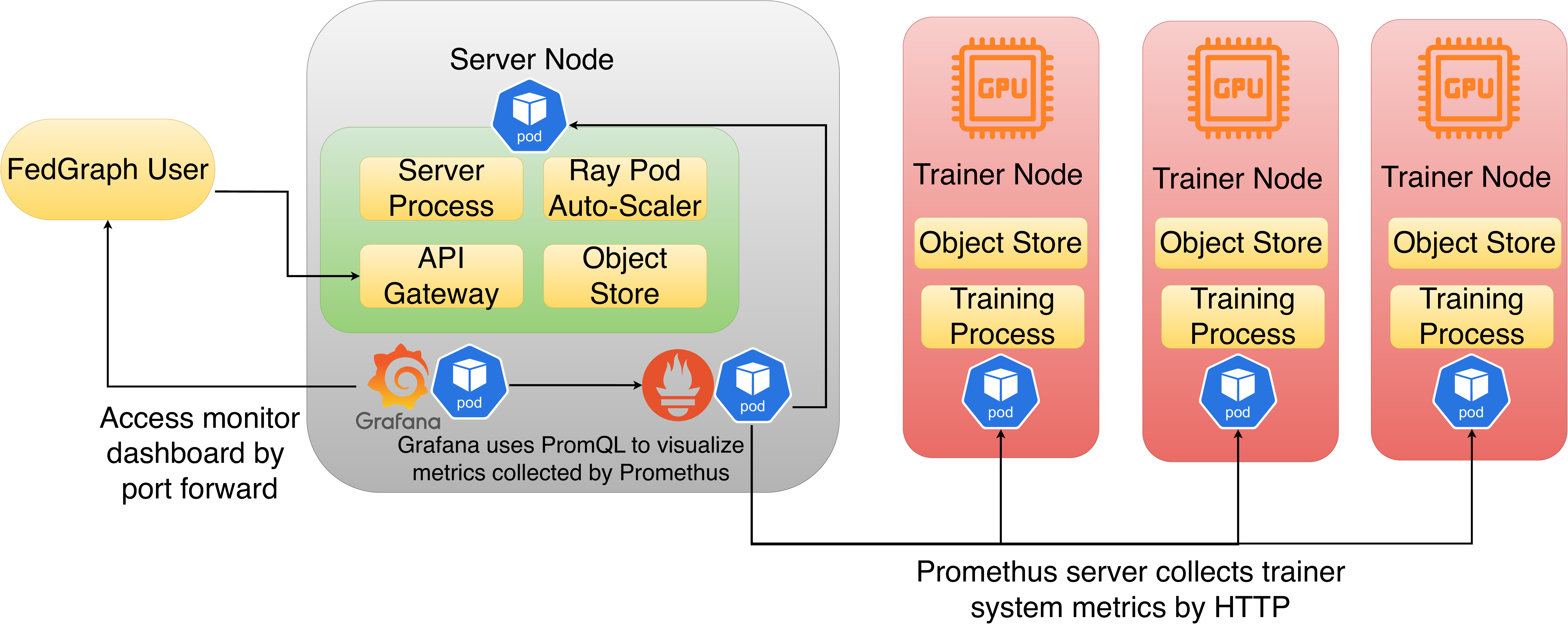}
    \caption{FedGraph Monitoring System Architecture.}
    \vspace{-5mm}
    \label{fig:fedgraph_k8s_monitor_system}
\end{wrapfigure}
The FedGraph Monitoring System, illustrated in Figure~\ref{fig:fedgraph_k8s_monitor_system} (right), uses a Monitor Class to track key system metrics, including running time, CPU/GPU utilization, memory consumption, and communication costs between the server and clients. Examples of the monitoring dashboard are provided in Figure~\ref{fig:fedgraph_nc_curves}.

\textbf{Communication Cost Logging}: The system records data transfer rates between the FedGraph User and the API Gateway to assess network performance and identify potential bottlenecks.\\
\textbf{Training Time and Accuracy Logging}: The monitoring framework captures the training duration and model accuracy throughout the learning process, enabling comparisons across configurations. \\
\textbf{CPU/GPU and Memory Usage Tracking}: Resource consumption across components, such as the Server Process, Ray Pod AutoScaler, API Gateway, and Object Store, is continuously monitored to provide insights into processing efficiency and to detect memory bottlenecks or potential leaks. 

Section~\ref{sec:benchmark-tasks} further illustrates FedGraph's resource usage profiles under various experimental settings.



\begin{wrapfigure}{r}{0.47\textwidth}
    \centering
    \vspace{-14.5mm}
    \includegraphics[width=0.47\textwidth]{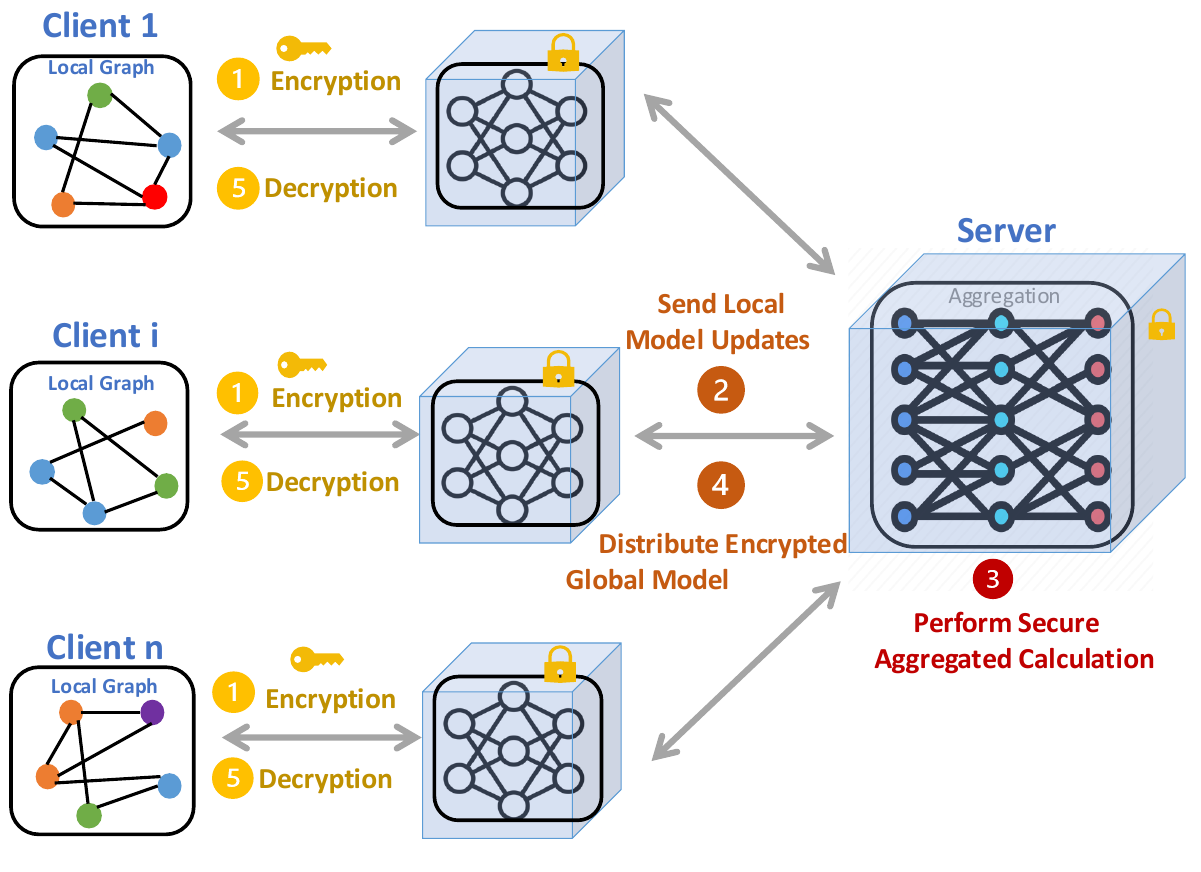}
    \caption{FedGraph Homomorphic Encryption for Secure Aggregation on Encrypted Data.}
    \vspace{-10mm}
    \label{fig:fedgraph_he_system}
\end{wrapfigure}
\subsection{FedGraph Homomorphic Encryption}
FedGraph supports homomorphic encryption (HE) to enable end-to-end secure computation, ensuring user privacy during training (Figure~\ref{fig:fedgraph_he_system}). To our knowledge, it is the first FGL library to incorporate HE as a core capability. While prior work has applied HE to specific FGL scenarios~\citep{yao2024fedgcn}, FedGraph generalizes this support by addressing two key requirements of realistic FGL settings: (i) secure feature aggregation prior to training~\citep{yao2024fedgcn}, and (ii) secure model aggregation during training~\citep{zhang2024deep,kim2025subgraph}.

\textbf{Pre-Training Aggregation}: For algorithms such as FedGCN that require feature aggregation before training, FedGraph applies HE to protect data privacy. Each client encrypts its local node features and transmits the encrypted data to the server. The server performs aggregation directly on the ciphertext and returns the result, which clients then decrypt to obtain the aggregated neighbor features without knowing the original feature.

\textbf{Federated Aggregation During Training}: Each client first encrypts its model update before transmission. The central server then performs homomorphic aggregation on the encrypted updates to compute the global model without accessing any plaintext data (Figure~\ref{fig:fedgraph_he_system})~\citep{zhang2020batchcrypt,jin2023fedml}.









\begin{wrapfigure}[10]{r}{0.52\textwidth}
    \centering
    \vspace{-0.27in}
    \includegraphics[width=0.52\textwidth]{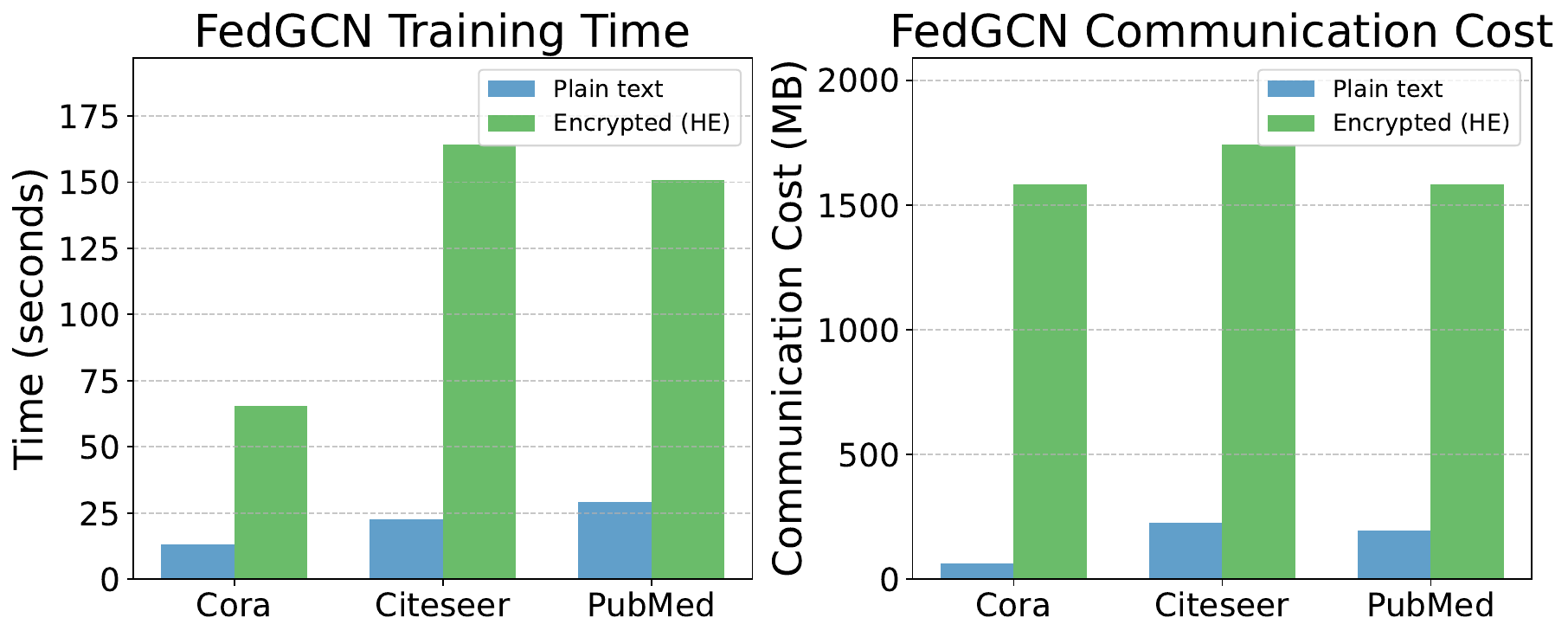}
    \caption{Comparison of FedGCN performance on plaintext (blue) and with HE encryption (green).}
    \label{fig:fedgcn_he_compare}
\end{wrapfigure}

\textbf{Homomorphic Encryption Overhead}: Figure \ref{fig:fedgcn_he_compare} compares the training time and communication cost for the FedGCN method with and without homomorphic encryption.
Since FedGCN involves both pre-training and training-phase aggregation, it serves as a representative case for evaluating the overhead introduced by HE across the full pipeline. As shown in the figure, HE significantly increases communication overhead, particularly during the pre-training phase, where feature matrices are often substantially larger than model parameters. These findings motivate our exploration of communication-efficient techniques in Section~\ref{sec:case}. Detailed benchmarks, including HE configurations and performance breakdowns, are in Appendix~\ref{sec:appendix-he}.


\subsection{FedGraph Kubernetes}
\begin{wrapfigure}[16]{r}{0.65\textwidth}
    \centering
    \vspace{-15mm}
    \includegraphics[width=0.65\textwidth]{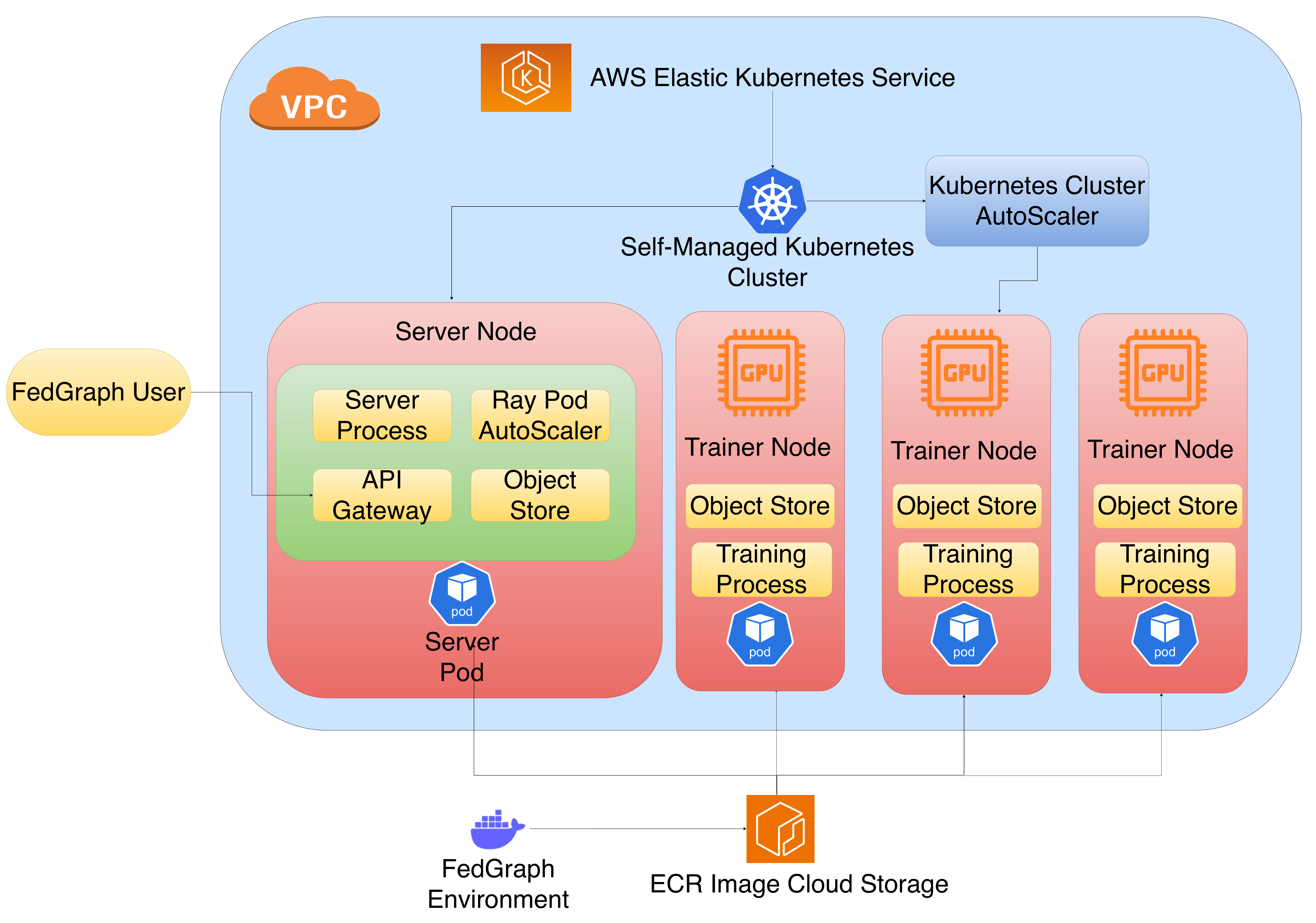}
    \caption{Kubernetes Service for FedGraph Cluster Management.}
    \label{fig:fedgraph_k8s}
\end{wrapfigure}
To support diverse training workloads and scale to a large number of clients, FedGraph employs Kubernetes as the foundation of its infrastructure layer. Unlike prior platforms (Table~\ref{tab:former_work}) that lack native Kubernetes integration, FedGraph is designed to scale efficiently to both large client populations and massive graph datasets. As shown in Figure~\ref{fig:fedgraph_k8s}, FedGraph leverages Amazon Web Services Elastic Kubernetes Service to enable flexible and dynamic resource management. A self-managed Kubernetes cluster equipped with GPU instances is deployed to handle computationally intensive tasks. The architecture consists of a master node responsible for orchestration, task scheduling, resource allocation, and cluster health monitoring, while worker nodes execute distributed training processes in parallel. 
To enhance scalability and cost-efficiency, we integrate the Kubernetes Cluster Autoscaler, which dynamically adjusts worker nodes based on workload demands. This ensures resources are used only when needed, maintaining high efficiency under varying system loads.


\subsection{Optimizations for Scalability}
FedGraph is designed to support large-scale federated learning, handling graphs with thousands to millions of nodes through configurable training features.

\textbf{Client Selection for Large-Scale Federated Learning}: FedGraph uses a selective client engagement mechanism in Appendix~\ref{sec:client_selection}, allowing the user to specify a fraction of clients that send the model update at each training round~\citep{li2019convergence}. This approach reduces communication costs at the server and resource usage at the clients, enabling scalability to thousands of clients.

\textbf{Minibatch Training for Federated Updates}:
Each client processes only a subset of its local graph, reducing computation and memory demands. This also enables devices with limited resources relative to the size of their training datasets to participate in the training, improving convergence speed without overwhelming network or client resources.

\textbf{Communication and Resource Optimization}:
FedGraph optimizes communication, minimizing data transfer between clients and the server. Kubernetes dynamically manages Server Pods and Trainer Pods to meet workload demands, ensuring efficient resource usage as the system scales.

These features collectively enable FedGraph to support federated learning at a massive scale while maintaining resource efficiency and performance.





\section{Case Study: Communication and Computation Efficient Federated Node Classification with Low Rank Feature Compression}\label{sec:case}

Efficient communication is a critical challenge in federated graph learning, especially when handling large graphs or deploying privacy-preserving methods such as homomorphic encryption. To address this, we implement low-rank feature compression within FedGraph that significantly reduces communication overhead while preserving model accuracy.

\subsection{Architecture Support in FedGraph}
FedGraph's modular design enables the seamless integration of low-rank compression. It separates the pre-training feature aggregation and model training phases, allowing different optimization strategies at each stage. In this case study, we apply HE in both pre-training and model training aggregations, while the low rank compression is applied during pre-training. The HE interface, integrated across multiple algorithms, naturally supports low-rank encrypted aggregation due to its additive structure.
\subsection{Low Rank Method for Pre-Train Feature Aggregation}
In FedGCN, pre-training communication involves aggregating feature information across clients for nodes with cross-client edges, which may be encrypted and securely aggregated in order to preserve privacy~\citep{yao2024fedgcn}.
We implement a low-rank method for pre-train feature aggregation using client-side projection of feature information to reduce communication cost. 
The server first generates a random projection matrix $\mathbf{P} \in \mathbb{R}^{d \times k}$ where $k \ll d$ is the specified rank (e.g., $k = 100$). The server then distributes $\mathbf{P}$ to all clients. Each client $i$ computes the projected feature matrix $\mathbf{\hat{X}}_i = \mathbf{X}_i \mathbf{P}$ where $\mathbf{\hat{X}}_i \in \mathbb{R}^{n_i \times k}$ with $n_i$ nodes and then sends $\mathbf{\hat{X}}_i$ to the server. The server aggregates these low-rank projections $\mathbf{\hat{X}}_{agg} = \sum_{i=1}^{m} \mathbf{\hat{X}}_i$, where $m$ is the number of clients, and distributes the result back to clients. To prevent potential reconstruction of another client's original features, the projection matrix $\mathbf{P}$ can also be encrypted before being distributed. This adds a layer of protection against inversion attacks on the shared aggregated features. This approach significantly reduces communication overhead in both directions due to only communicating low-rank information. Since our HE interface supports addition on encrypted data, it can safely perform the aggregation procedure on encrypted projected features, preserving client privacy.

\subsection {Performance Evaluation}
\begin{figure*}[ht]
    \centering
    \includegraphics[width=0.47\textwidth]{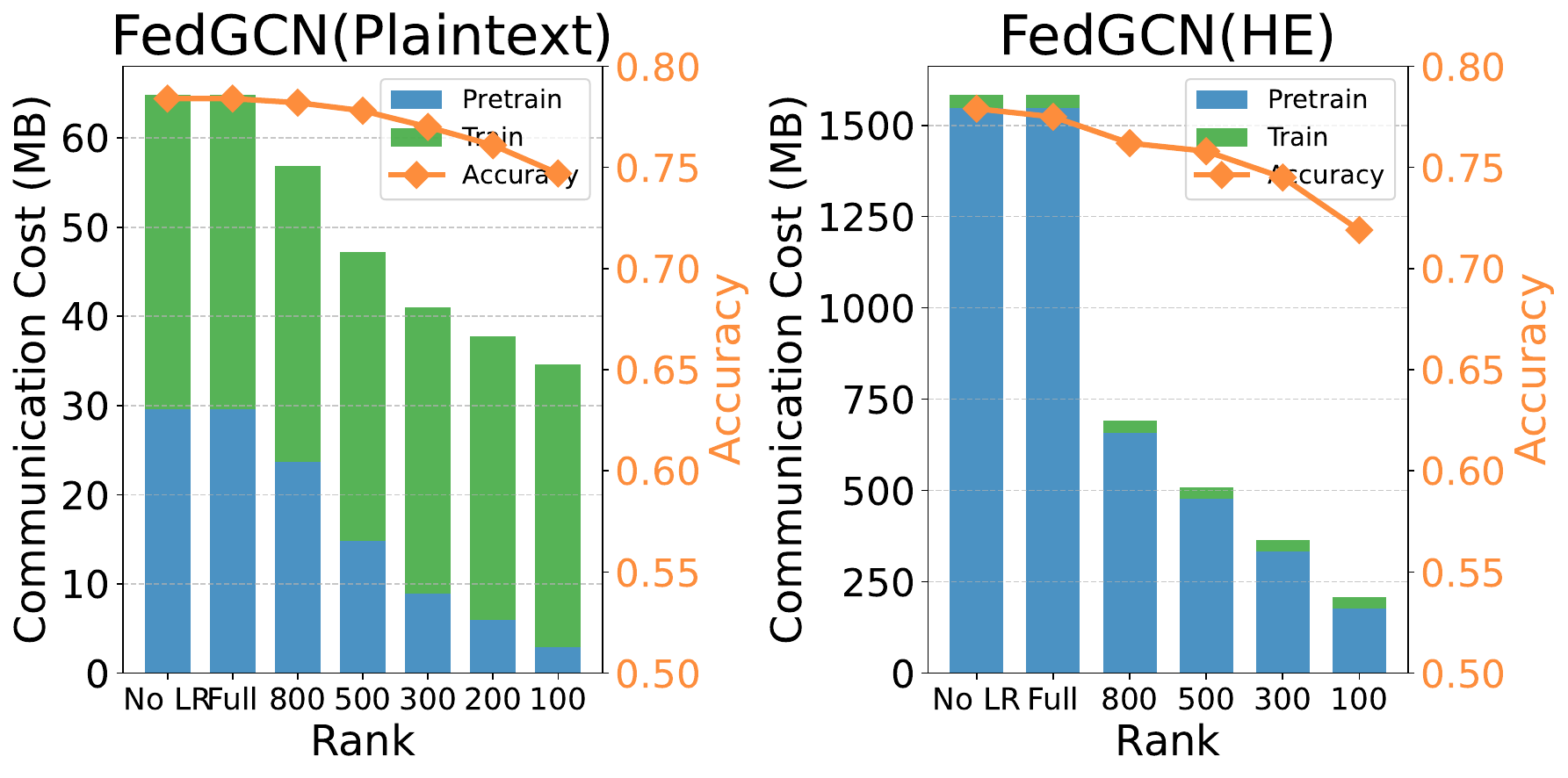}
    \includegraphics[width=0.47\textwidth]{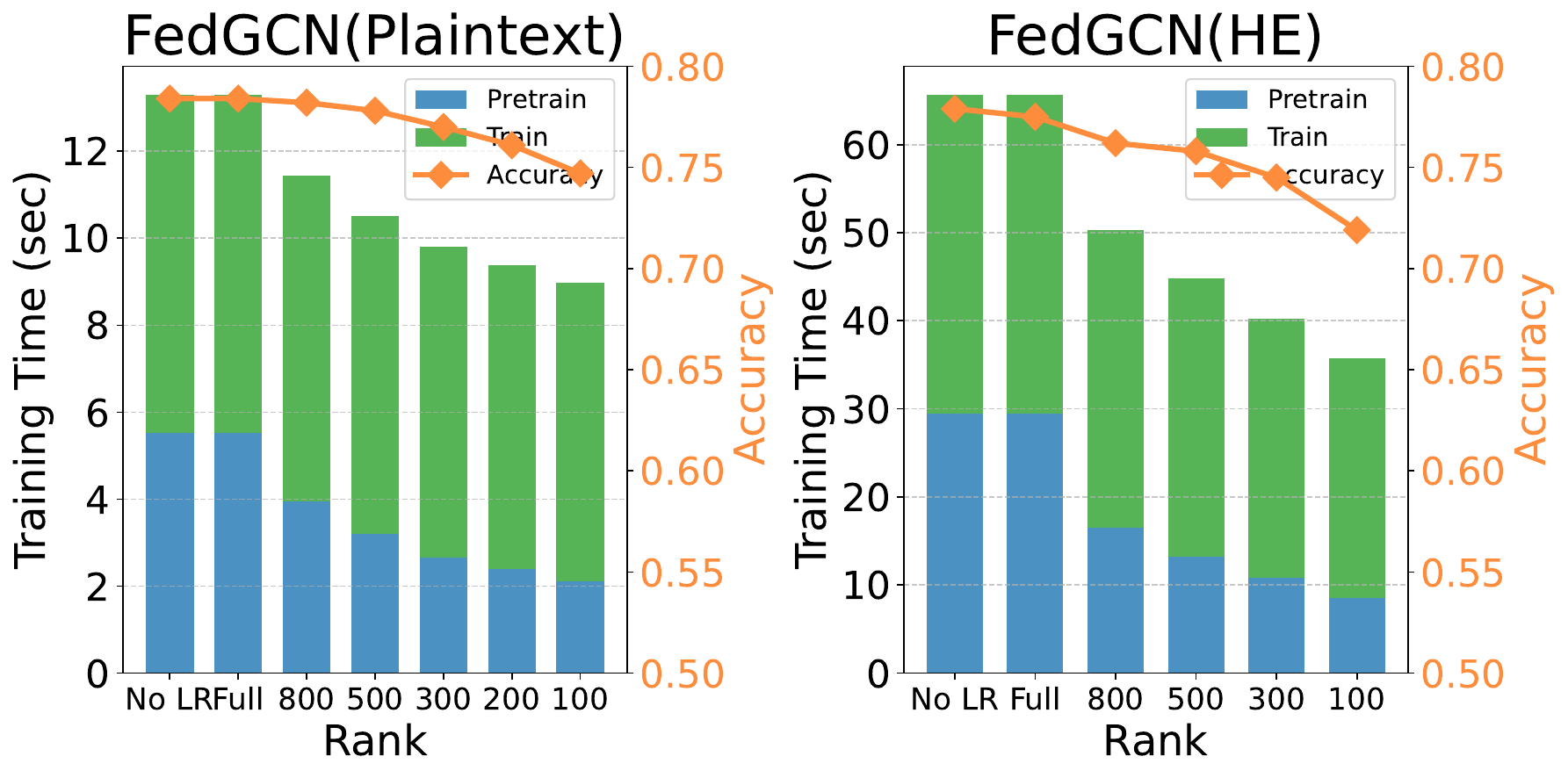}
    \caption{Comparison of communication cost (first and second plots) and training time (third and fourth plots) for FedGCN with different low-rank compression settings. We show results with plaintext and with HE. Each bar represents the total cost divided into pre-training (blue) and training (green) phases, with accuracy plotted as an orange line to show performance trade-offs.}
    \label{fig:fedgcn_low_rank_compare}
\end{figure*}

We evaluate the impact of low-rank compression on communication cost, training time, and accuracy in FedGCN using the Cora dataset. Compression ranks range from the full 1433-dimensional features to rank 100, yielding up to 93\% reduction. As shown in Figure~\ref{fig:fedgcn_low_rank_compare}, HE alone introduces high communication overhead, especially in pre-training, but this is substantially mitigated by applying low-rank projection. Accuracy remains stable even as rank decreases, demonstrating the effectiveness of low-rank compression for communication-efficient and privacy-preserving FGL.




\section{Benchmarking FedGraph on Graph Learning Tasks and Scalability}\label{sec:benchmark-tasks}
In this section, we benchmark FedGraph's performance on graph classification, node classification, and link prediction tasks. We then benchmark FedGraph's performance as we scale the size of the dataset, demonstrating the library's ability to scale well. Appendix~\ref{sec:appendix-experiments} provides additional results.




\subsection{Benchmarking Federated Graph Learning Tasks}
We first benchmark FedGraph across three representative graph learning tasks, evaluating accuracy, training time, and communication cost to provide a view of the system and model performance.

\subsubsection{Federated Graph Classification}
\begin{figure*}[htbp]
    \centering
    \begin{minipage}[b]{0.325\textwidth}
        \centering
        \includegraphics[width=\linewidth]{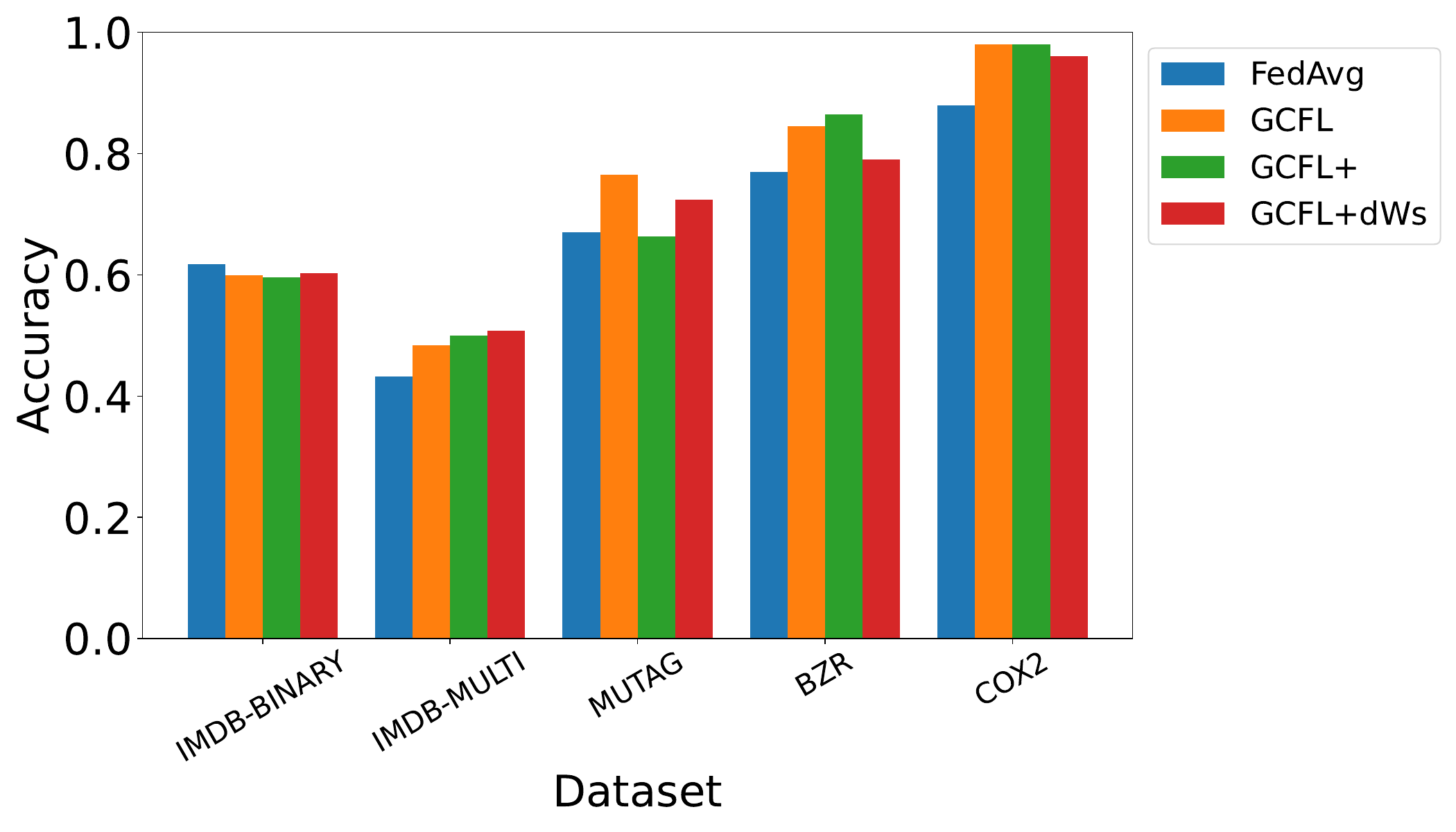}
    \end{minipage}
    \hfill
    \begin{minipage}[b]{0.325\textwidth}
        \centering
        \includegraphics[width=\linewidth]{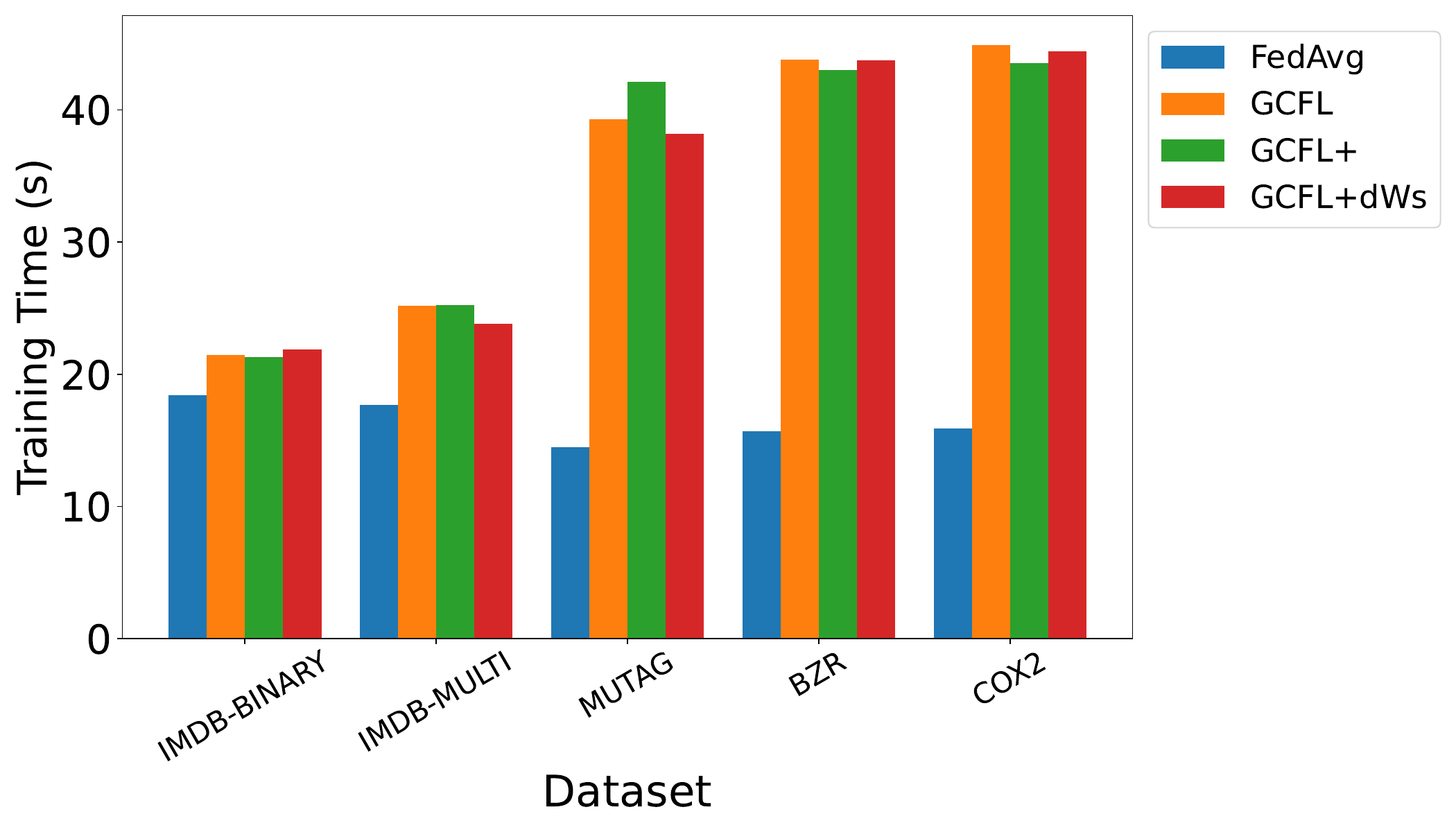}
    \end{minipage}
    \hfill
    \begin{minipage}[b]{0.325\textwidth}
        \centering
        \includegraphics[width=\linewidth]{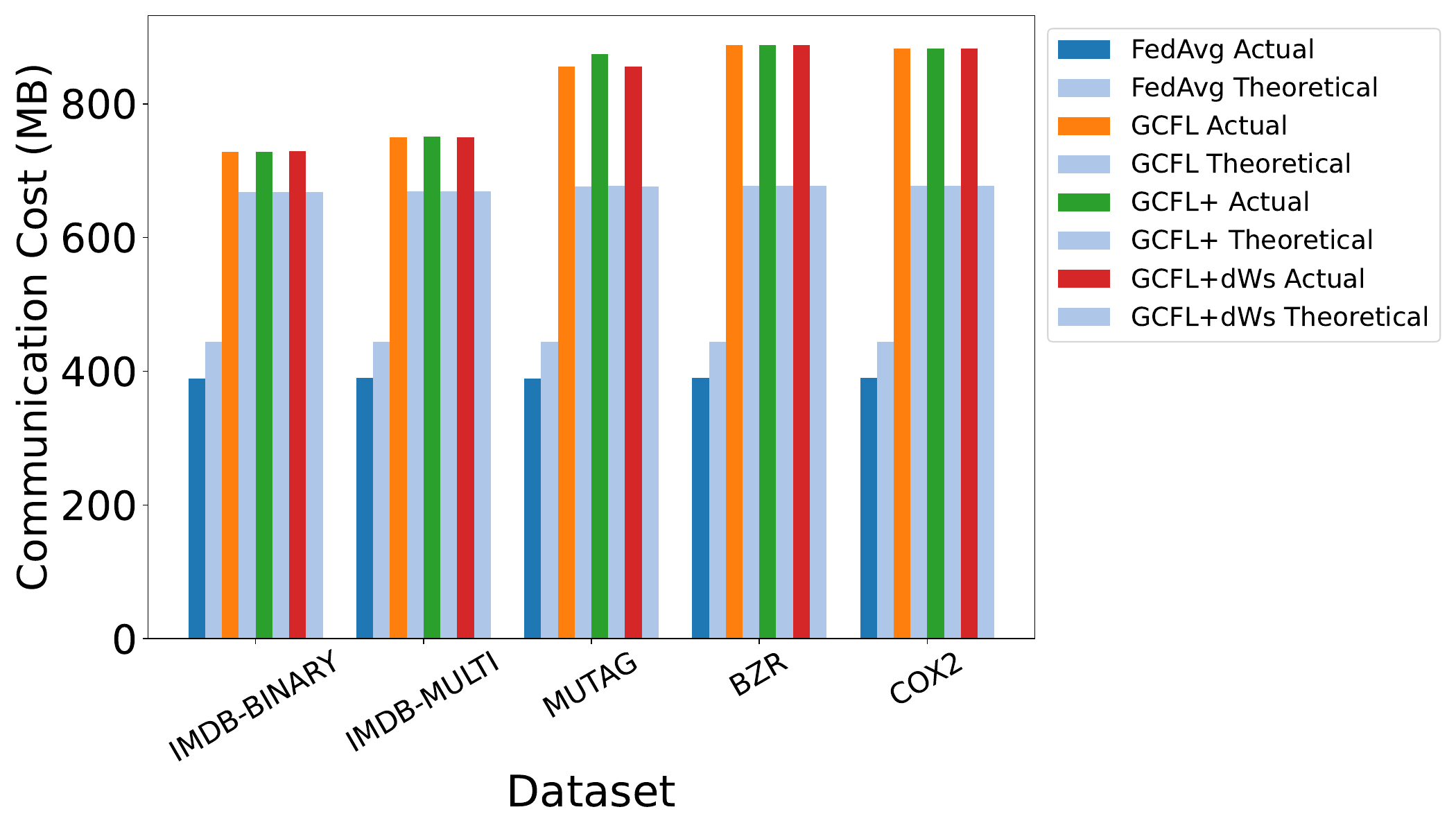}
    \end{minipage}
    \caption{Accuracy (left), Training Time (middle), and Communication Cost (right) Comparison across Federated Graph Classification Algorithms (10 clients). }
    \label{fig:fedgraph_gc_acc}
\end{figure*}

The graph classification experiment evaluates the performance of various federated learning algorithms on five benchmark datasets: IMDB-BINARY, IMDB-MULTI, MUTAG, BZR, and COX2. These datasets span diverse graph structures, including social and molecular networks, providing a comprehensive testbed for assessing algorithm effectiveness. 
As shown in Figure~\ref{fig:fedgraph_gc_acc}, the evaluation spans 200 training rounds across five datasets. GCFL+ and GCFL+dWs consistently achieve the highest accuracy, particularly on BZR and COX2. However, as illustrated in the middle and right plots, these gains come at the cost of significantly higher training time and communication overhead, especially on complex datasets like IMDB, reflecting their greater computational and communication demands. In contrast, FedAvg offers the lowest communication cost and the shortest, most consistent training time across all datasets, making it a practical choice for latency- and bandwidth-constrained environments.
\subsubsection{Federated Node Classification}
\begin{figure*}[htbp]
    \centering
    \begin{minipage}[b]{0.325\textwidth}
        \centering
        \includegraphics[width=\linewidth]{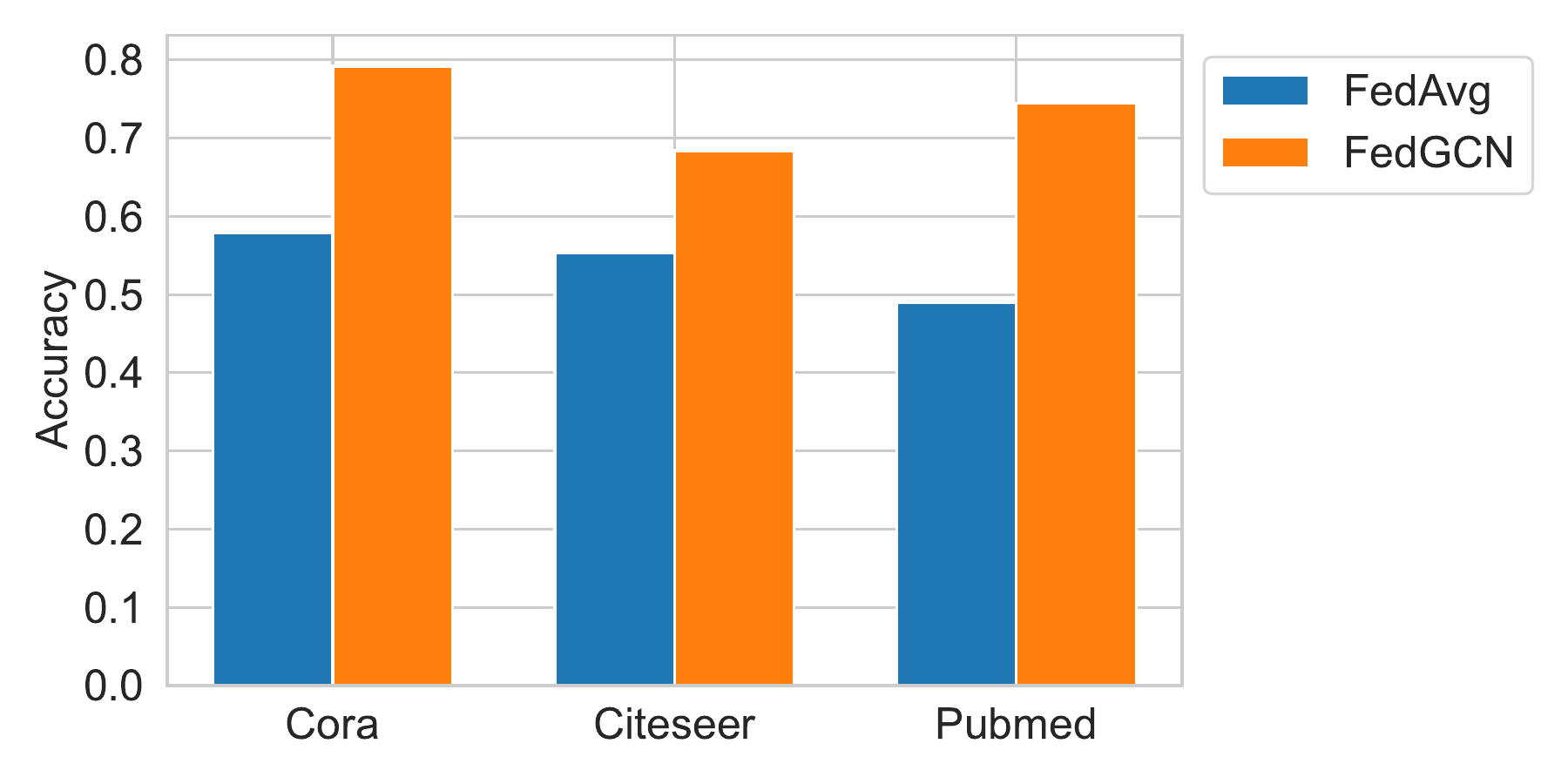}
    \end{minipage}
    \hfill
    \begin{minipage}[b]{0.325\textwidth}
        \centering
        \includegraphics[width=\linewidth]{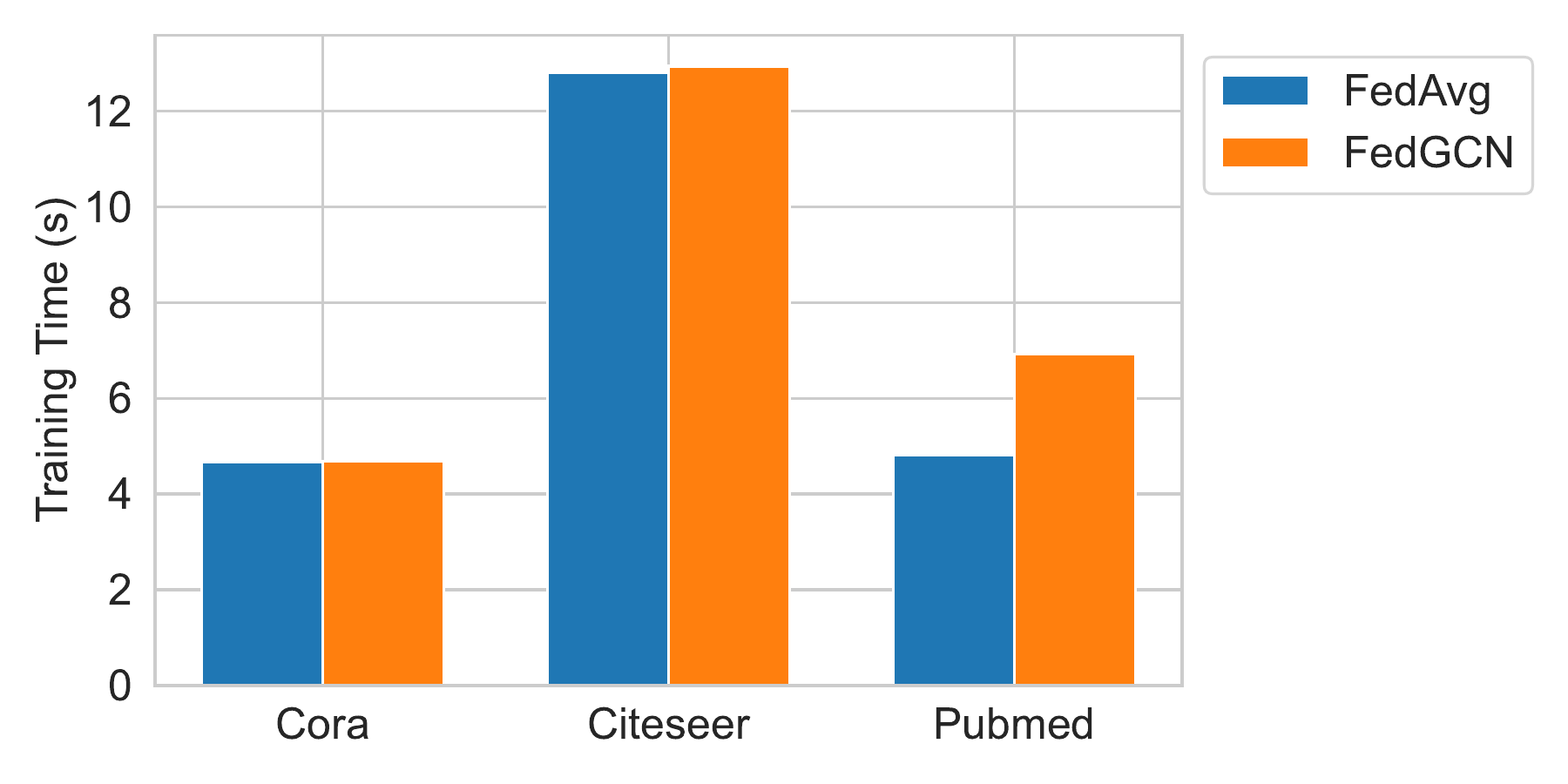}
    \end{minipage}
    \hfill
    \begin{minipage}[b]{0.325\textwidth}
        \centering
        \includegraphics[width=\linewidth]{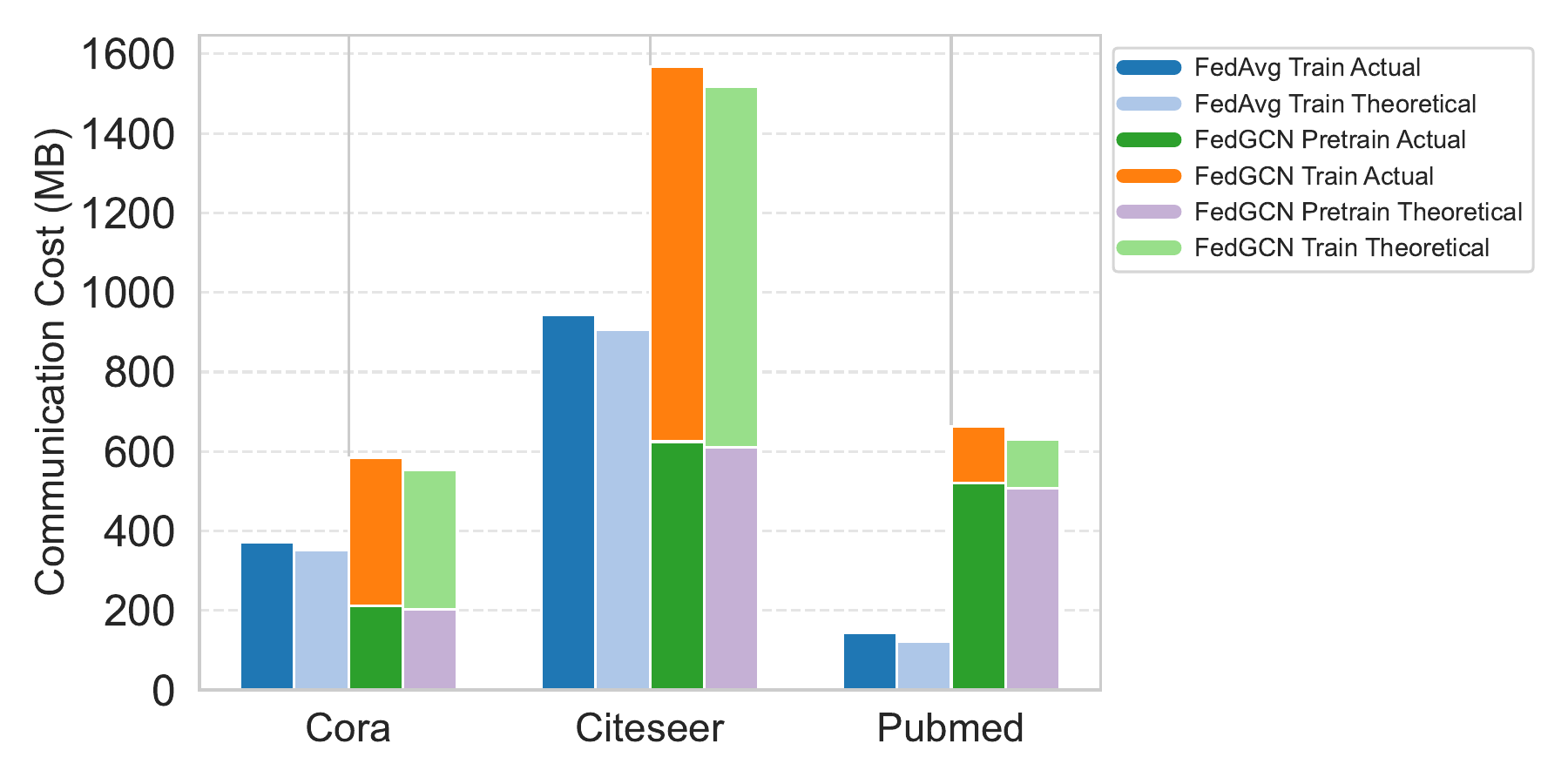}
    \end{minipage}
    \caption{Accuracy (left), Training Time (middle), and Communication Cost (right) Comparison across Federated Node Classification Algorithms under $\beta=10000$ (IID data distribution).}
    \label{fig:fedgraph_nc_acc}
\end{figure*}

We evaluate FedAvg and FedGCN on four node classification benchmarks: Cora, Citeseer, and PubMed. As shown in Figure~\ref{fig:fedgraph_nc_acc} (left), FedGCN consistently achieves higher accuracy than FedAvg across all datasets. The communication cost breakdown (Figure~\ref{fig:fedgraph_nc_acc}, right) reveals the source of this overhead: Compared to FedAvg, which does not have a pre-training round, FedGCN still has high pre-training communication costs due to aggregation of feature sums across clients. Notably, the observed communication costs closely match the theoretical values, validating the efficiency modeling of these algorithms. Such observation inspires the design of the new low rank algorithm in Section~\ref{sec:case}.




\subsubsection{Federated Link Prediction}
\begin{figure*}[htbp]
    \centering
    \begin{minipage}[b]{0.325\textwidth}
        \centering
        \includegraphics[width=\linewidth]{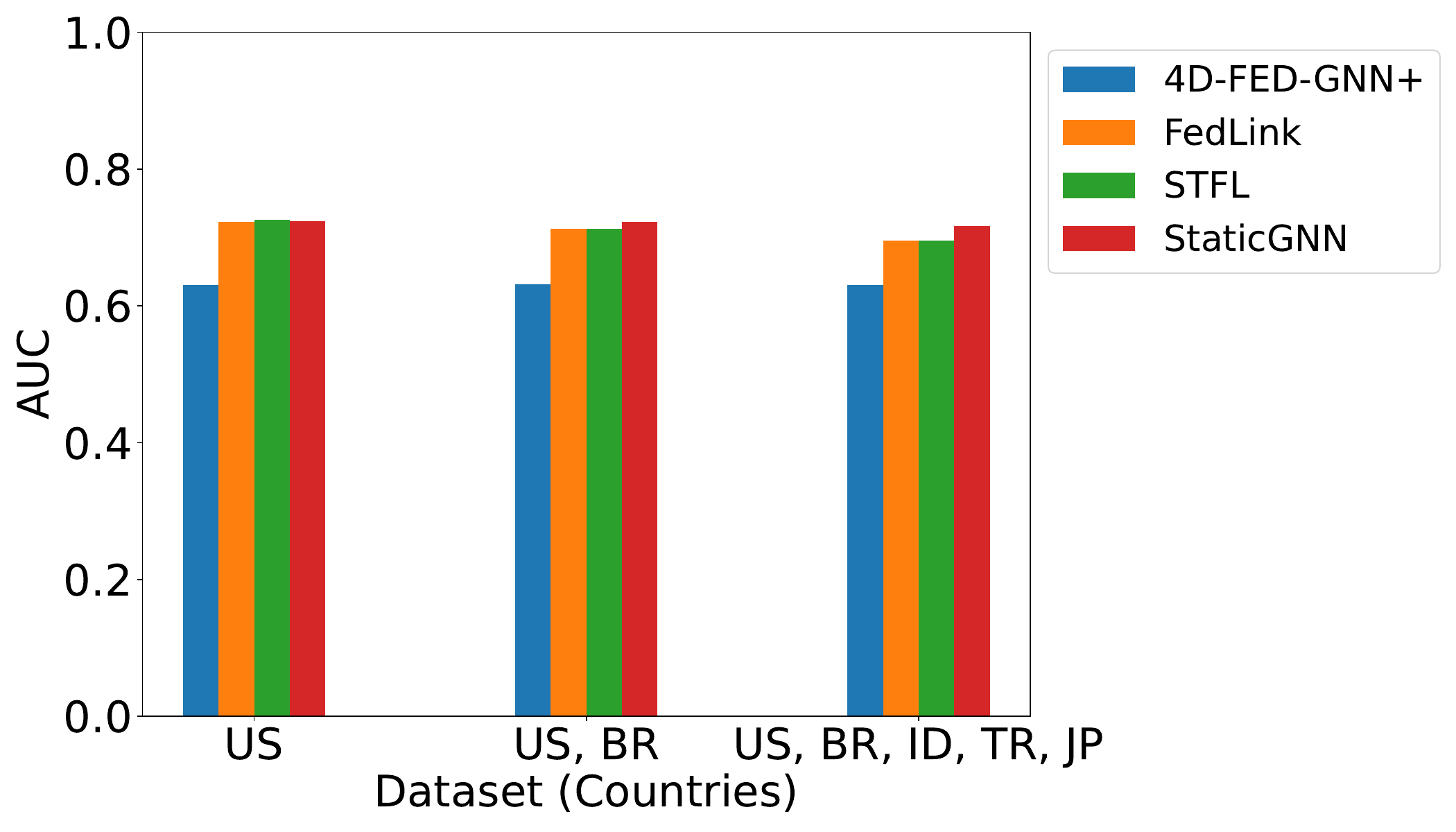}
    \end{minipage}
    \hfill
    \begin{minipage}[b]{0.325\textwidth}
        \centering
        \includegraphics[width=\linewidth]{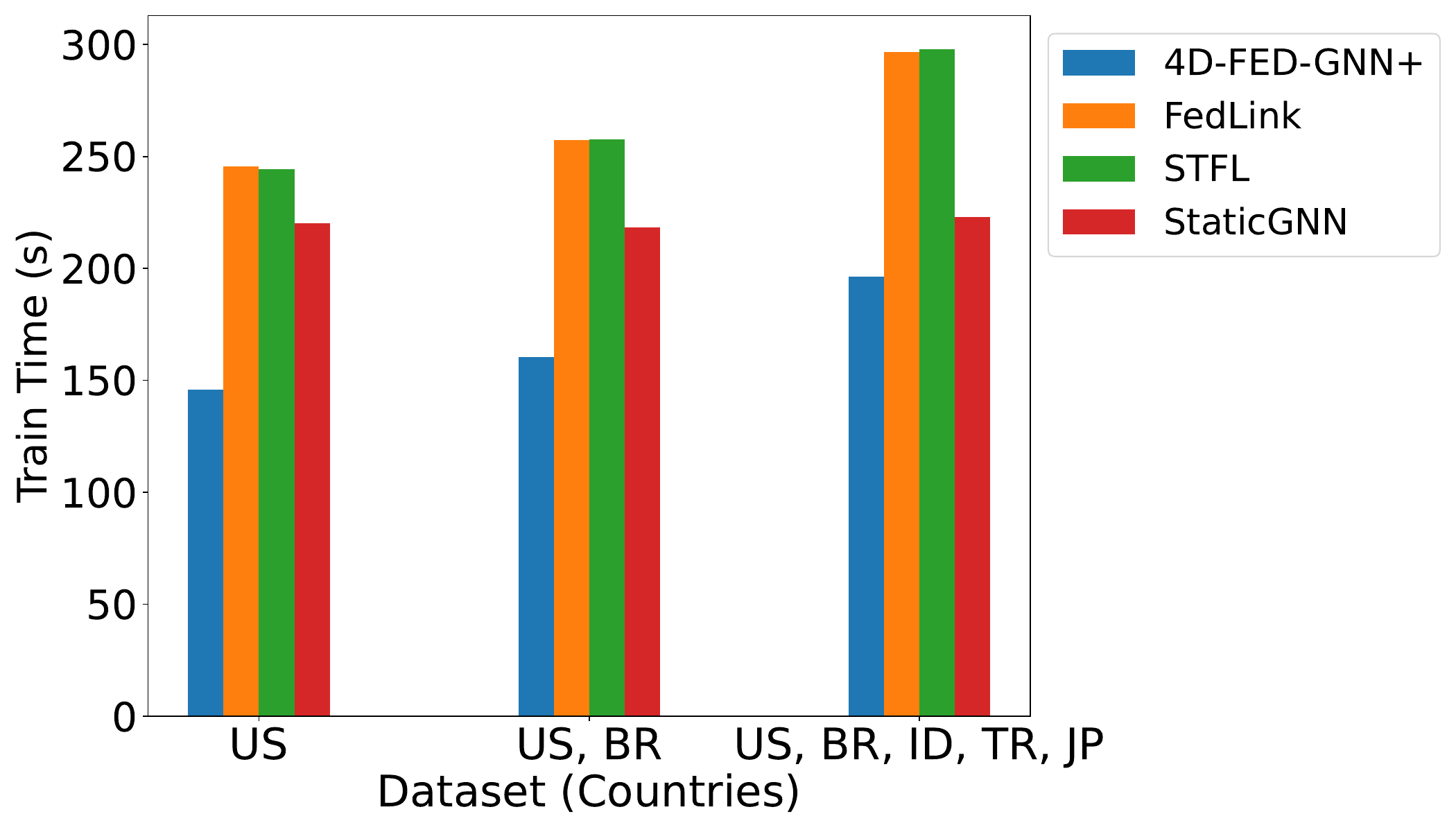}
    \end{minipage}
    \hfill
    \begin{minipage}[b]{0.325\textwidth}
        \centering
        \includegraphics[width=\linewidth]{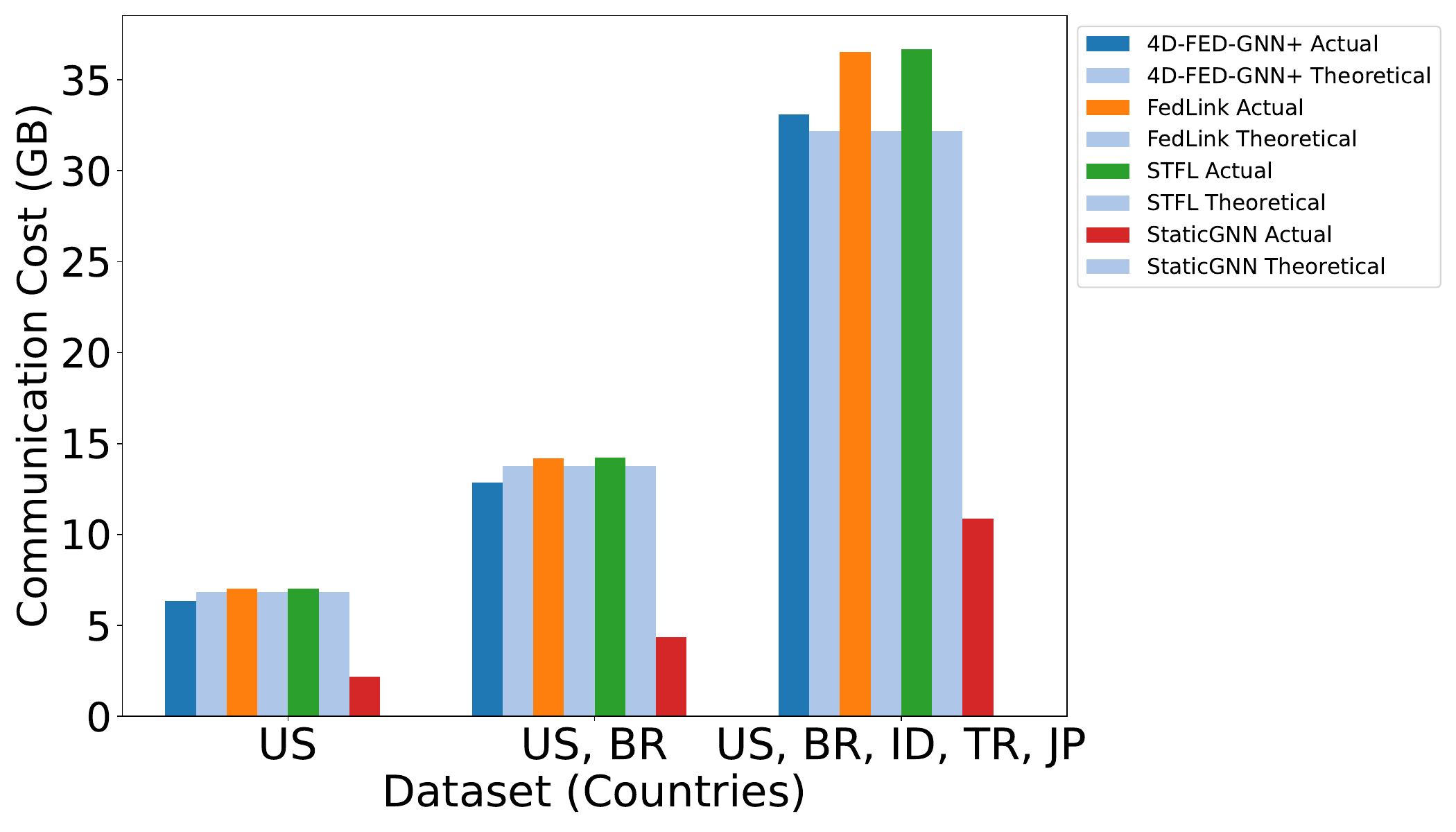}
    \end{minipage}
    \caption{AUC (left), Training Time (middle), and Communication Cost (right) Comparison across Federated Link Prediction Algorithms (10 clients).}
    \label{fig:fedgraph_lp}
\end{figure*}

In the federated link prediction setting, we simulate a scenario where each client holds region-specific data from the Foursquare Global-scale Check-in Dataset~\citep{yang2016participatory}, covering three geographic configurations: US; US and BR; and US, BR, ID, TR, and JP. This setup respects data privacy by avoiding raw data sharing across regions. We evaluate four algorithms, 4D-FED-GNN+, FedLink, STFL, and StaticGNN, across three key metrics: AUC for predictive accuracy, training time for computational efficiency, and communication cost for network efficiency. 

As shown in Figure~\ref{fig:fedgraph_lp}, FedLink and STFL achieve the highest AUC scores across all datasets, while StaticGNN and 4D-FED-GNN+ perform moderately well, but show lower AUC on the simpler US-only dataset. In terms of training time, FedLink and STFL incur the highest costs, particularly on the largest regional dataset. In contrast, 4D-FED-GNN+ demonstrates the shortest training times, suggesting suitability for fast iterative training. For communication efficiency, FedLink incurs the highest overhead, especially on large datasets, while StaticGNN consistently exhibits the lowest communication cost, making it the most network-efficient among the evaluated methods.





\subsection{System Performance Monitoring and Resource Utilization}
\begin{figure*}[ht]
    \centering
    \begin{minipage}[b]{0.29\textwidth}
        \centering
        \includegraphics[width=\linewidth]{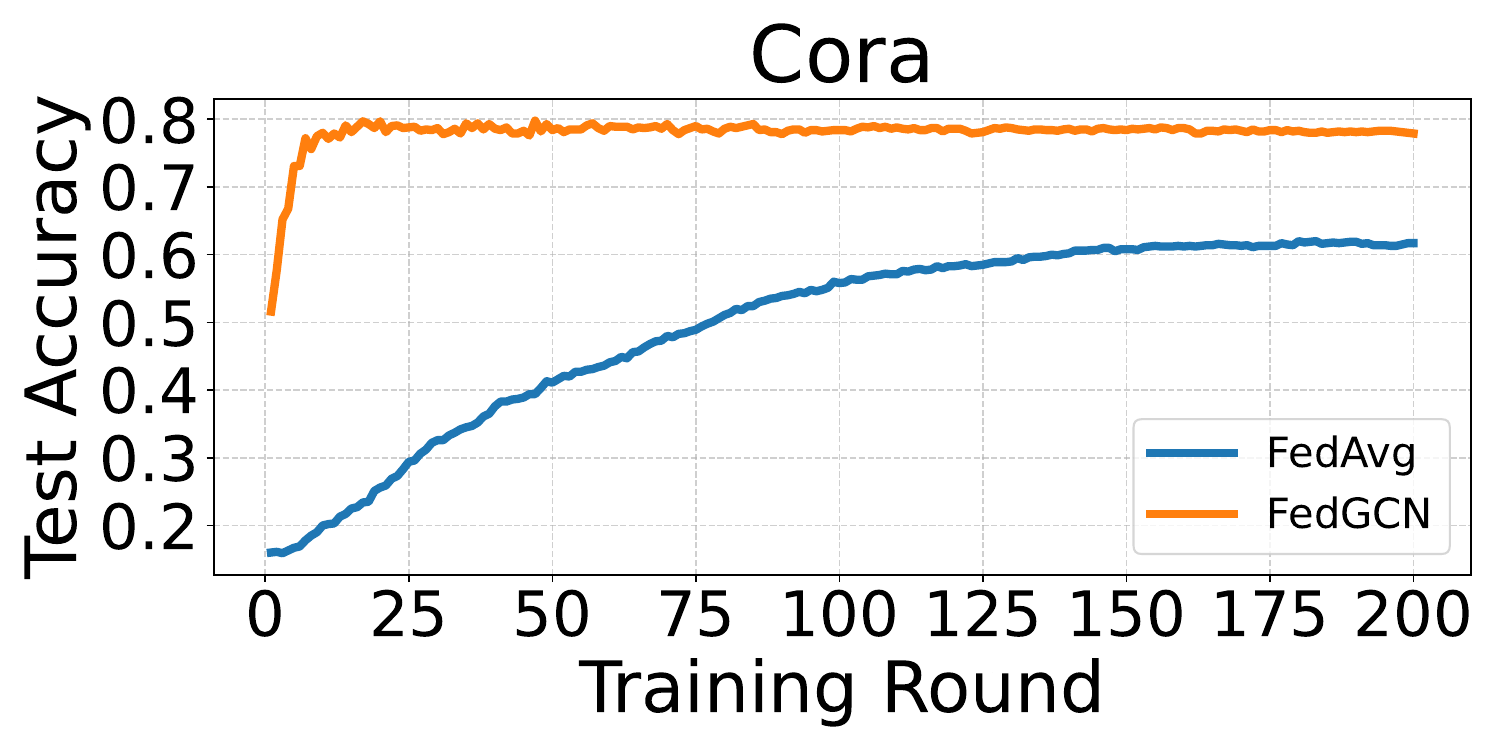}\\
        \includegraphics[width=\linewidth]{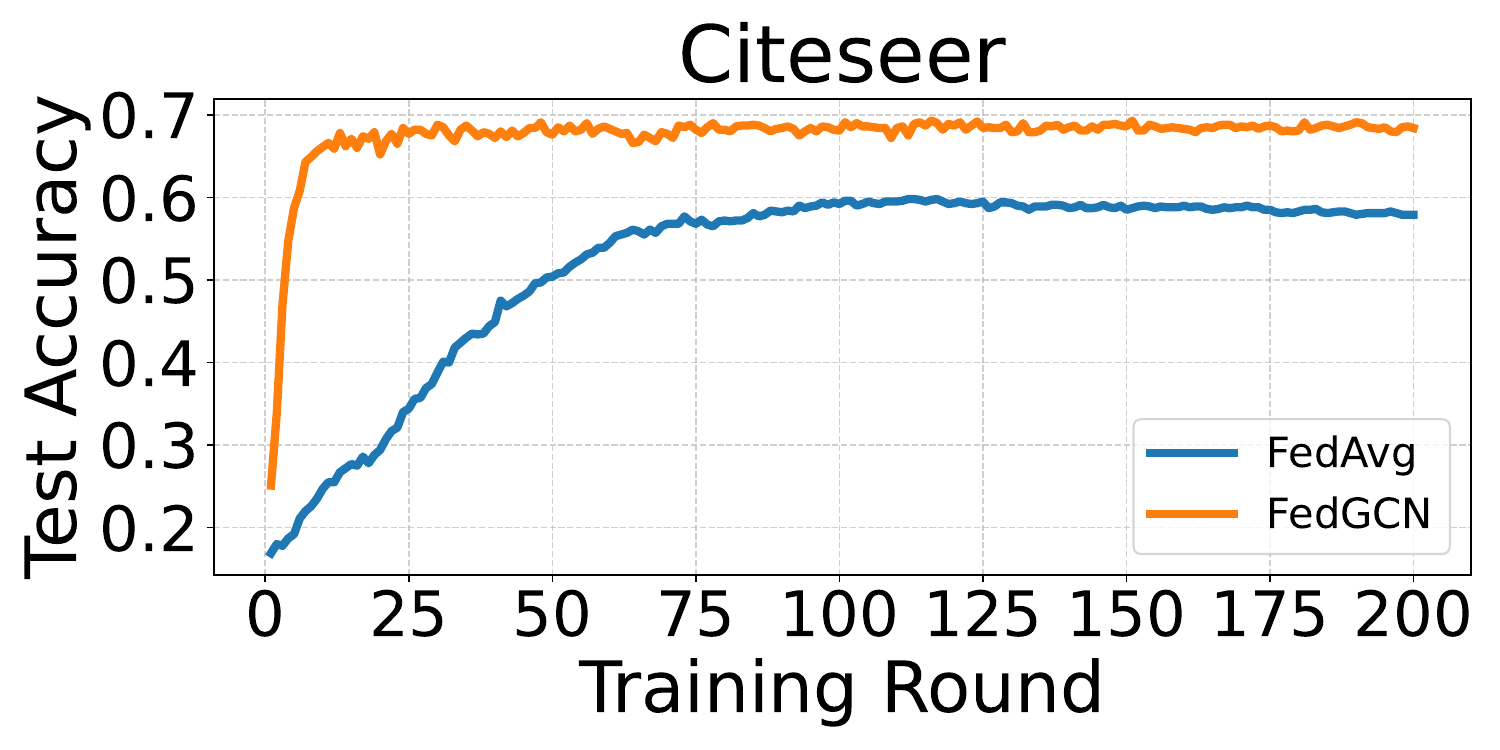}\\
        \includegraphics[width=\linewidth]{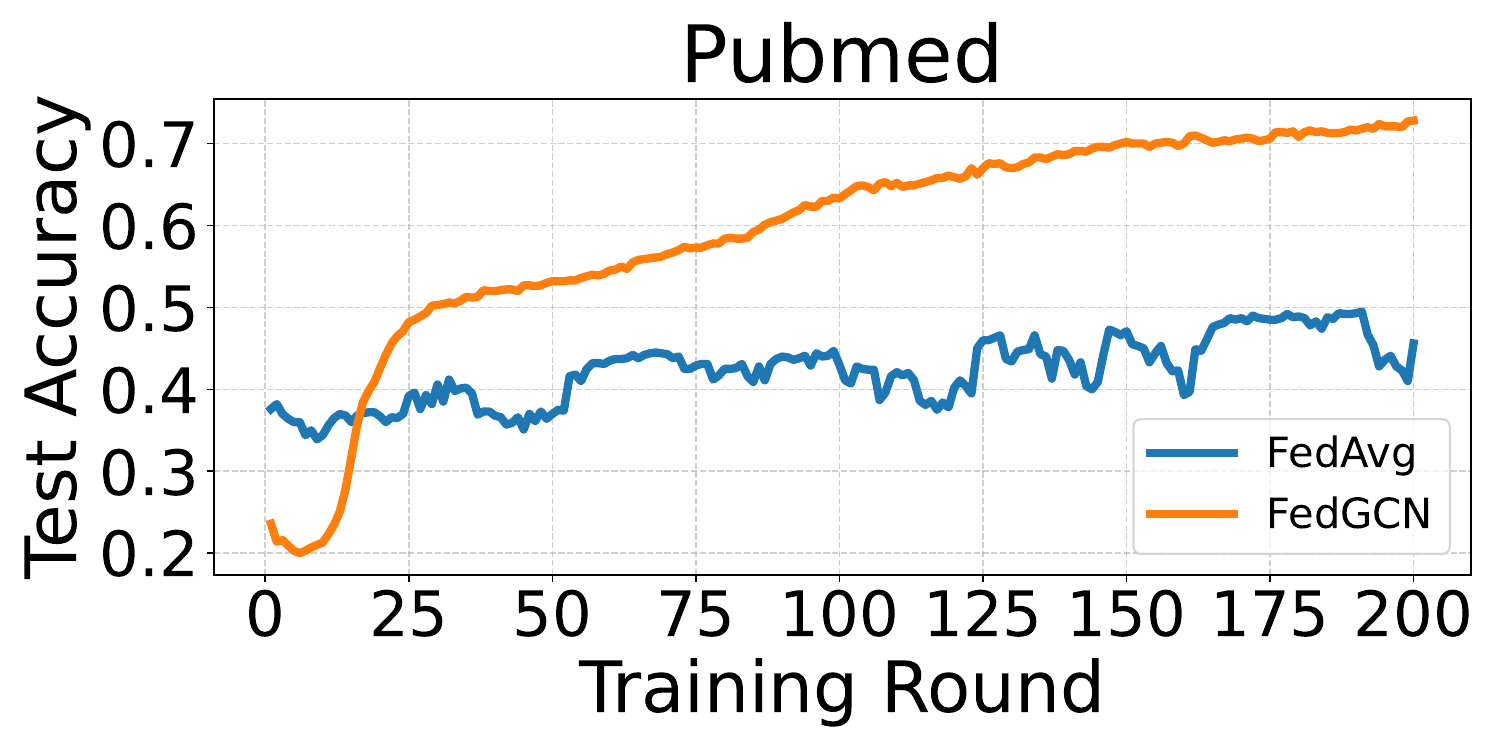}
    \end{minipage}
    \hfill
    \begin{minipage}[b]{0.7\textwidth}
        \centering
        \includegraphics[width=\linewidth]{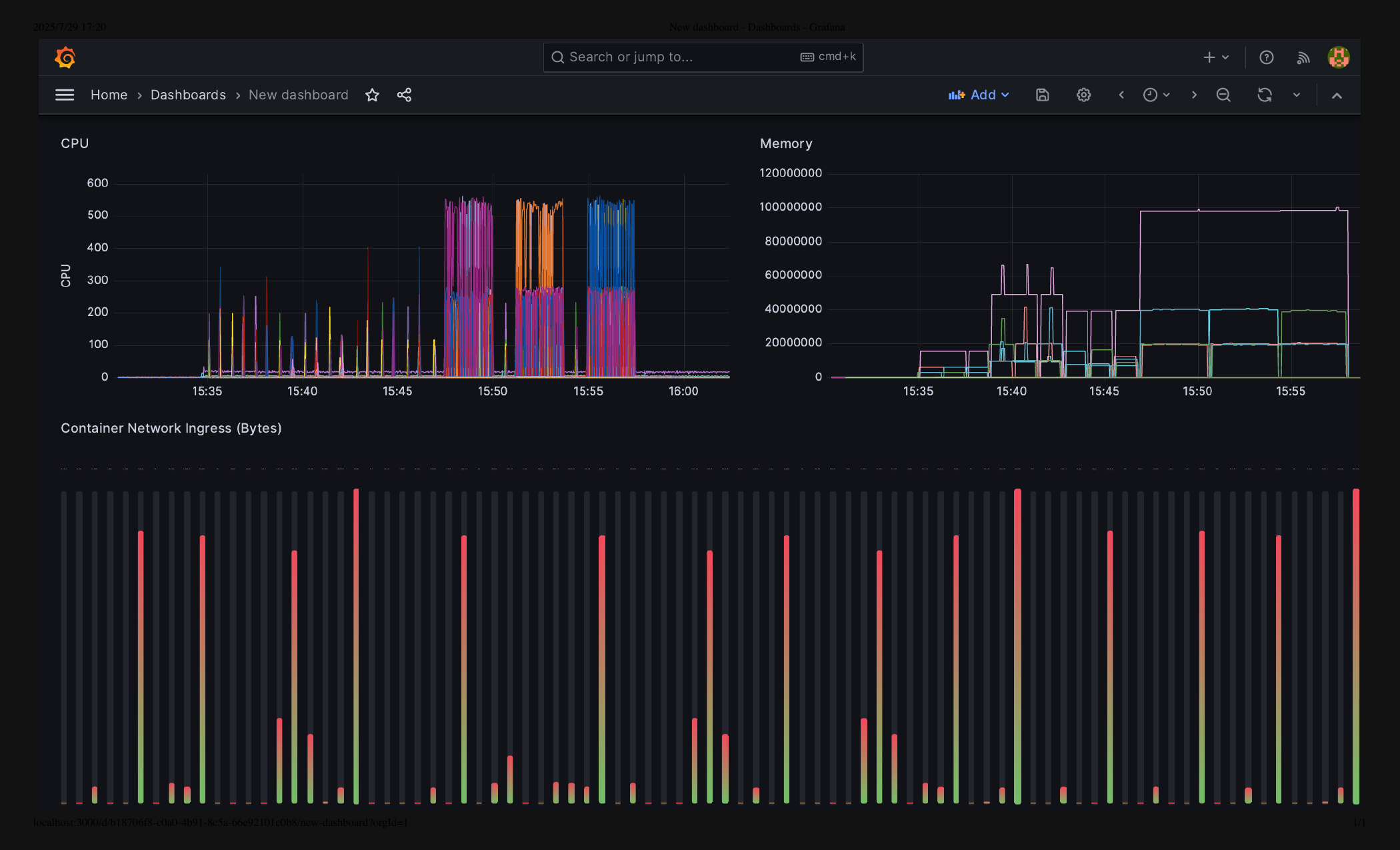}
    \end{minipage}
    \caption{Test Accuracy across Global Training Rounds (left, Cora/Citeseer/Pubmed) and Grafana Dashboard Showing CPU, Memory, and Network usage (right).}
    \label{fig:fedgraph_nc_curves}
\end{figure*}

We monitor system performance across 10 training nodes and a central server node using Grafana, which visualizes real-time metrics collected via Prometheus. Figure~\ref{fig:fedgraph_nc_curves} provides an overview of system behavior during node classification experiments. The first three plots show global test accuracy across training rounds for the Cora, Citeseer, and Pubmed datasets. FedGCN demonstrates significantly faster convergence and higher final accuracy than FedAvg across all datasets. Figure~\ref{fig:fedgraph_nc_curves} (right) shows CPU, memory, and network usage over time, captured via Grafana. Lighter workloads, such as Cora and Citeseer, result in lower and less frequent CPU utilization spikes. In contrast, Pubmed and Ogbn-Arxiv induce higher and more sustained CPU usage due to their larger graph sizes and increased communication overhead. The pattern of usage spikes aligns with scheduled training rounds, reflecting real-time computational demands at scale.

\subsection{Real-World Dataset with Realistic Client Data Distribution}

\begin{figure}[ht]
    \centering
    \includegraphics[width=0.36\textwidth]{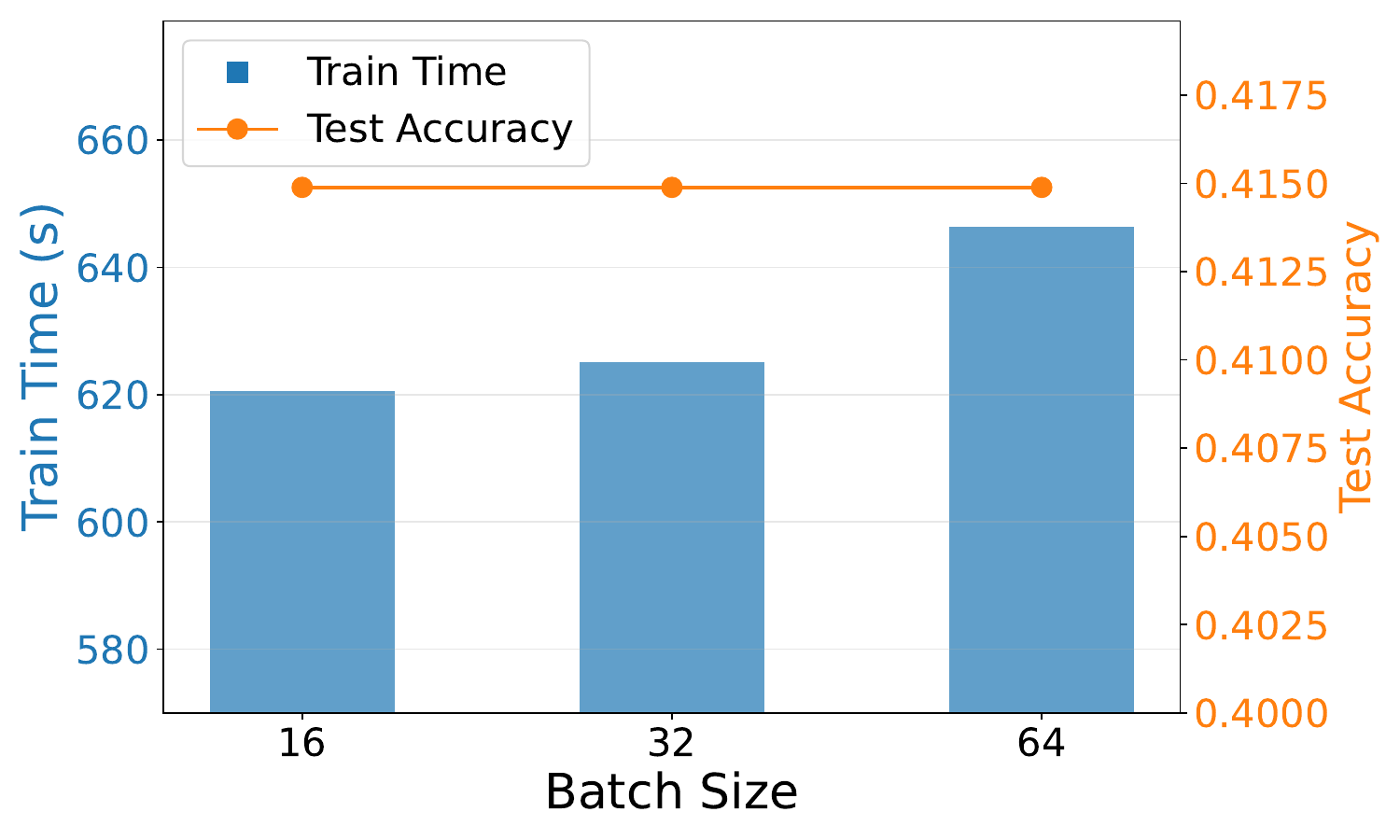}
    \hspace{1mm}
    \includegraphics[width=0.6\textwidth]{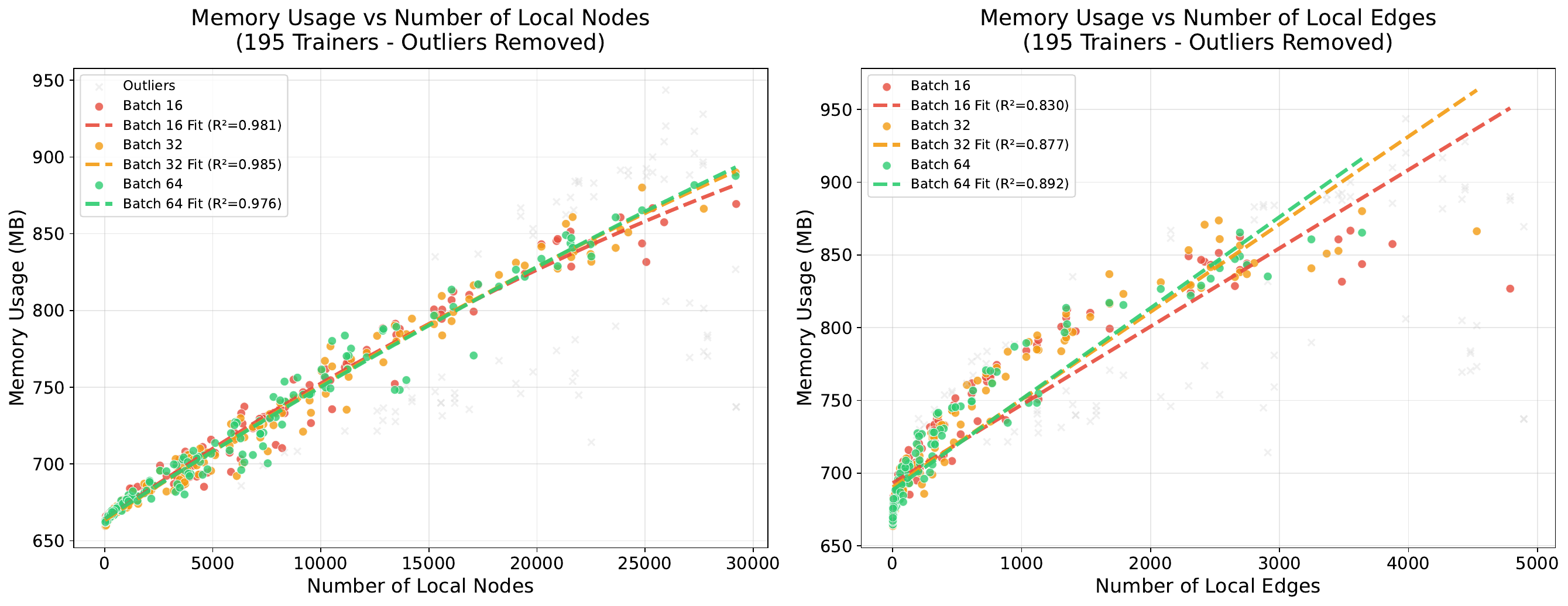}
    \caption{Training Time, Test Accuracy, and Memory Usage for each client on Ogbn-Papers100M under Different Batch Sizes (800 rounds).}
    \label{fig:fedgraph_100M_traintime}
\end{figure}
We finally evaluate FedGraph at scale on the Ogbn-Papers100M dataset (over 50GB), one of the largest publicly available graph benchmarks. We use Hugging Face for dataset storage and partitioning, with the Trainer class managing local data loading for each client. The data is distributed across 195 clients, with node counts assigned following a power law distribution based on country population sizes, mimicking realistic federated environments where larger clients hold more data. 

As shown in Figure~\ref{fig:fedgraph_100M_traintime}, we assess the effect of varying batch sizes (16, 32, and 64) on training time and test accuracy over 800 training rounds. Training time increases modestly with batch size due to additional per-round computation. Test accuracy remains nearly unchanged, with a slight gain from batch size 16 to 32 and a plateau at 64, suggesting flexibility in batch size selection without compromising model performance. Memory usage remains stable at approximately 17.5 GB, demonstrating FedGraph's scalability and efficient resource handling in large-scale settings.

\subsection{Benchmarking Scalability under Increasing Clients}
We next evaluate the scalability of FedGraph by varying the number of clients in Table~\ref{tab:scalability}. As the number of clients increases, the overall communication cost grows substantially, eventually becoming the primary bottleneck. In contrast, the training time per client decreases since each client processes a smaller subgraph. 

\begin{table}[ht]
\centering
\caption{Training and communication time (seconds) for datasets under varying client numbers.}
\begin{tabular}{
c
|S[table-format=3.2]S[table-format=3.2]
|S[table-format=3.2]S[table-format=3.2]
|S[table-format=3.2]S[table-format=3.2]
|S[table-format=3.2]S[table-format=3.2]}
\toprule
\multirow{2}{*}{Clients} &
\multicolumn{2}{c|}{Cora} &
\multicolumn{2}{c|}{CiteSeer} &
\multicolumn{2}{c|}{PubMed} &
\multicolumn{2}{c}{OGBN-arXiv} \\
\cmidrule(lr){2-9}
 & {Train} & {Comm} & {Train} & {Comm} & {Train} & {Comm} & {Train} & {Comm} \\
\midrule
5  & 1.39 & 1.69 & 1.58 & 2.78 & 2.08 & 1.55 & 127.71 & 4.48 \\
10 & 1.36 & 2.78 & 1.79 & 6.55 & 1.77 & 2.60 & 45.82  & 5.95 \\
15 & 1.56 & 3.99 & 2.40 & 9.58 & 1.57 & 3.84 & 21.77  & 7.77 \\
20 & 1.49 & 4.87 & 2.07 & 13.62& 1.83 & 4.63 & 17.89  & 9.24 \\
\bottomrule
\end{tabular}
\label{tab:scalability}
\end{table}


\section{Conclusion}\label{sec:conclusion}
In this paper, we presented FedGraph, a Python library designed for benchmarking federated graph learning algorithms. Unlike general federated learning platforms, FedGraph supports a diverse set of algorithms and enables systematic comparisons across algorithms, datasets, and system configurations. It features fully distributed training, homomorphic encryption for privacy-preserving scenarios, and a built-in system profiler to measure communication and computation overhead. The modular API allows easy integration of custom datasets and algorithms. Through extensive experiments, including low-rank compression and large-scale training on graphs with up to 100 million nodes, we demonstrate that FedGraph is a practical and scalable tool for real-world FGL evaluation.

While this work focuses on enabling privacy-preserving federated graph learning, future efforts are needed to explore more robust privacy risk assessments, additional optimization strategies, and the inclusion of a broader range of FGL algorithms to expand benchmark coverage and better support industrial deployment.

\newpage
\bibliography{ref}
\bibliographystyle{unsrtnat}



\newpage
\appendix
\section{Frequently Asked Questions}

\subsection{Client Selection}\label{sec:client_selection}

FedGraph supports two client selection methods that randomly or uniformly select clients at each round, as in server\_class.py. Server-side algorithm components can also be added by modifying the server class.

\begin{lstlisting}
assert 0 < sample_ratio <= 1, "Sample ratio must be between 0 and 1"

num_samples = int(self.num_of_trainers *ample_ratio)

if sampling_type == "random":
    selected_trainers_indices = random.sample(
        range(self.num_of_trainers), num_samples
    )
elif sampling_type == "uniform":
    selected_trainers_indices = [
        (
            i
            + int(self.num_of_trainers * sample_ratio)
            * current_global_epoch
        )
        % self.num_of_trainers
        for i in range(num_samples)
    ]

else:
    raise ValueError("sampling_type must be either 'random' or 'uniform'")
\end{lstlisting}

\subsection{Easy Integration with New Baselines}
We agree that adding more FGL methods to FedGraph could make it more useful to users. Indeed, our main goal is to provide a library for benchmarking the real system performance of federated graph learning methods. Though we believe we have covered most state-of-the-art methods, we acknowledge that some methods from the literature are missing, so we also make the library easy to add new methods for researchers. Researchers can create a new training class based on an existing training method in the library (e.g., FedAvg, FedGCN).

\subsection{Communication Overhead after HE with Low Rank}

Homomorphic Encryption (HE) has sometimes significant drawbacks, in particular, high computing and communication loads. We include an HE implementation in the FedGraph framework as some prior work on federated graph learning has proposed HE as a way to preserve privacy, e.g., ~\cite{effendi2024privacy,fu2022federated,ni2021vertical}. Thus, researchers in the field may wish to evaluate the effects of HE on various graph learning algorithms, or even evaluate new ways to combine HE with federated graph learning. As a benchmark platform, we do not advocate for or against such ideas; our goal is simply to facilitate the evaluation of federated graph learning algorithms that require HE implementations. For example, researchers may wish to quantitatively compare the training and communication time of their algorithms with and without HE in order to precisely measure the overhead induced by using HE. We will clarify this point in the revised manuscript.

Our work on low-rank HE is meant to serve as an example case study of the types of research that FedGraph enables with HE. We fully agree that it does not reduce the overhead of communication during training. However, FedGraph also facilitates implementing low-rank HE schemes for the aggregation process, such as FedPara~\cite{hyeon2021fedpara}.

\subsection{Comparison with Other Framework}
The FedGraphnn and FederatedScopeGNN libraries have not been maintained since 2023. When running FedAvg, due to using the same graph training library, PyG, and the distributed setup, FedGraph has a similar run time as FedGraphnn and FederatedScopeGNN. As shown in Table~\ref {tab:former_work}, the main advantage is supporting new algorithms, homomorphic encryption, system-level profilers, and large-scale optimization like low rank, client selection, mini-batch, etc. Given the optimization methods (e.g., low rank in Figure~\ref{fig:fedgcn_low_rank_compare}), FedGraph can run much faster than these frameworks.

\subsection{Support Differential Privacy and Homomorphic Encryption}
FedGraph supports differential privacy(DP) for aggregation as an option in configuration. Our implementation of DP achieves comparable performance to both the plaintext version and HE without having an accuracy loss. Table~\ref{tab:dp_he} provides a comparison of different matrix running FedGCN with Cora using plaintext, HE, and DP. Results are averaged over 5 runs. Both HE and DP protect the pre-training or training communication without exposing the raw data at the server in different approaches. This meets our goal of presenting FedGraph to provide user with the flexibility of using and choosing the privacy mechanism that best fits their specific needs.  

\begin{table}[ht]
\centering
\begin{tabular}{l|c|c|c|c}
\toprule
Framework & Pre-train Communication (MB) & Pre-train Time (s) & Total Time (s) & Accuracy \\
\midrule
Plaintext & 56.61   & 4.91  & 12.08 & 0.793 \\
HE        & 1208.87 & 17.49 & 40.91 & 0.791 \\
DP        & 57.69   & 5.60  & 13.09 & 0.792 \\
\bottomrule
\end{tabular}
\caption{Comparison of privacy preservation methods in terms of pre-train communication cost, pre-train time, total time, and accuracy.}
\label{tab:dp_he}
\end{table}

\subsection{Why HE Affects Model Accuracy}

HE performance is highly sensitive to hyperparameter settings. For example, in Table~\ref{tab:supported_datasets} and Table~\ref{tab:supported_algorithm}, we vary the Polynomial Modulus Degree to balance security and computational capacity. A higher degree supports more complex computations but incurs greater overhead, while a lower degree restricts the allowable computation depth before noise dominates, leading to truncated or simplified operations that degrade model accuracy. In some cases, decryption itself becomes inaccurate, with the error magnitude determined by both the chosen hyperparameters and the encrypted values. Since the decrypted outputs are used for graph model training, such errors accumulate and ultimately reduce training accuracy.

\subsection{Run on Edge Devices}
In this paper, we focus on AWS cloud experiments as many of our envisioned use cases for federated graph learning--including learning on medical record data stored at different hospitals, or user-product consumption data stored in different countries--falls into the cross-silo federated learning paradigm, where clients are likely large servers or computers (hospitals or countries respectively, in the two examples above). FedGraph could also be run on Linux-based mobile or Internet of Things devices, e.g., Jetson Nanos, as might be appropriate for mobile applications like wearable health sensing.

\section{FedGraph Code Structure}\label{sec:appendix-code}

Fedgraph library can be separated into six modules:

\textbf{Data Process Module: data\_process.py.} This module is responsible for data generation and processing. It should be called before calling the runner, so it is only applicable for node classification and graph classification in the latest version. (In link prediction, the dataset is generated and processed inside the runner.)

\textbf{Runner Module: federated\_methods.py.} Task-based runners are defined in this module. For node classification and link prediction, there's a shared runner that could call inner modules to perform federated graph learning. For graph classification, it will further assign the program to an algorithm-based runner.

\textbf{Server Classes Module: server\_class.py.} It defines different task-based server classes including  server\_NC, server\_GC, and server\_LP.

\textbf{Trainer Classes Module: trainer\_class.py.} It defines different task-based trainer(client) classes including trainer\_NC, trainer\_GC, and trainer\_LP.

\textbf{Backbone Models Module: gnn\_models.py.} It defines different backbone model classes. Generally, it is task-based (i.e., each task corresponds to one backbone model), but for some models, their variants are also included, and the user could also switch the backbone model or define a new one themselves.

\textbf{Utility Functions: utils\_nc.py, utils\_gc.py, utils\_lp.py.} In the current version, the utility functions for different tasks are located in separate Python modules, which is convenient for development. In the later versions, it might be better to use the shared `utils.py`, and use different markers like “NC”, “GC”, and “LP” to distinguish them.

\section{FedGraph API and Runners}\label{sec:appendix-api}

\begin{figure}[ht]
    \centering
    \begin{subfigure}[b]{0.95\textwidth}
        \centering
        \includegraphics[width=\linewidth]{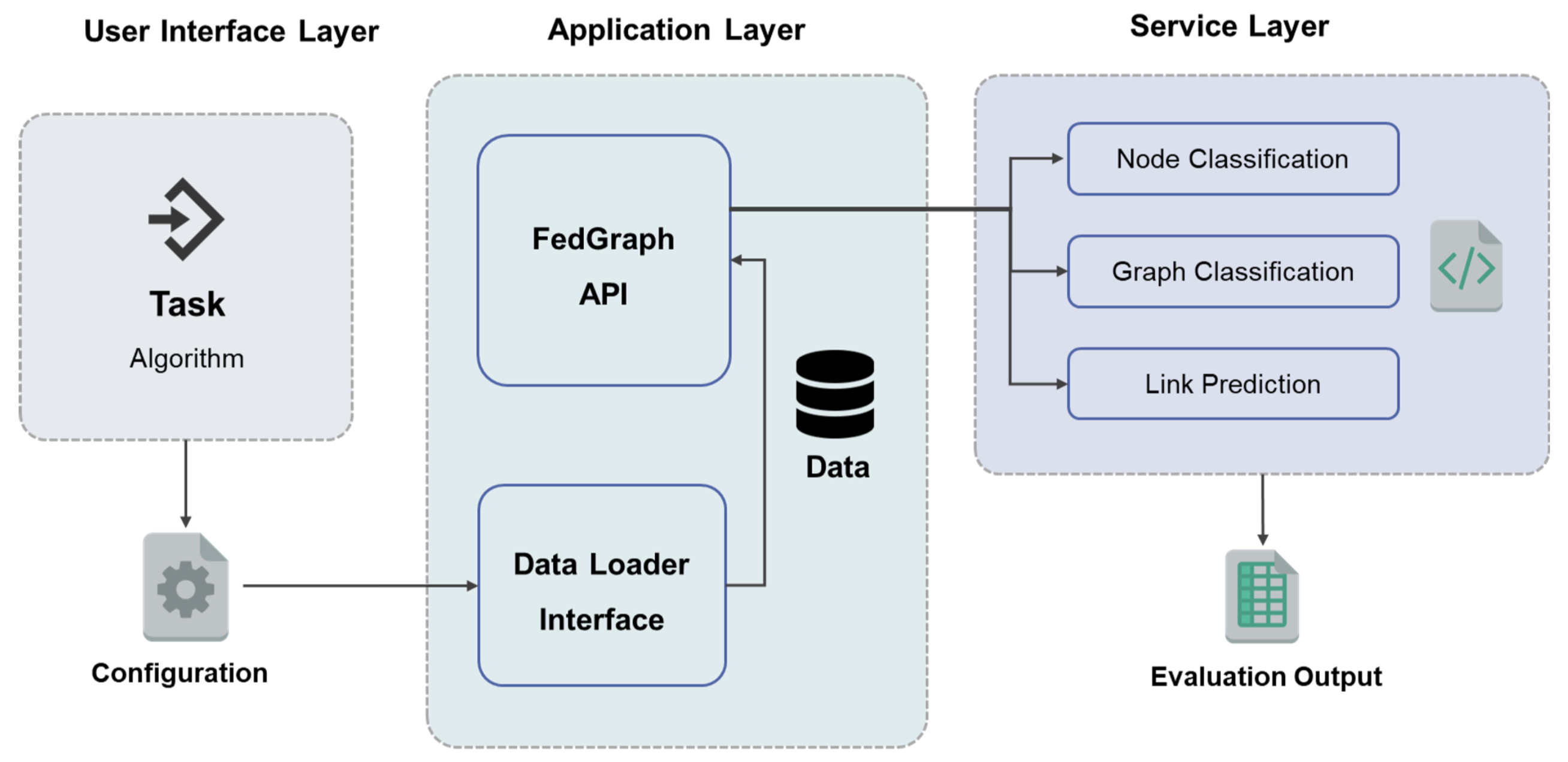}
    \end{subfigure}
    \caption{High-level Demonstration of FedGraph API Design}
    \label{fig:fedgraph_api}
\end{figure}

When calling the FedGraph API, as shown in Figure~\ref{fig:fedgraph_api}, Users can simply specify the name of the dataset and algorithm. The API will then call the corresponding data loader class to generate the required data, which is then automatically fed into the appropriate algorithm runner for tasks like Node Classification, Graph Classification, or Link Prediction. Additionally, users can seamlessly add their own datasets or federated graph learning algorithms if needed, as long as they satisfy the form requirements for the specified task.




The function run\_fedgraph is a general runner that receives the dataset and the configurations. It will further assign the program to a task-specified runner run\_NC , run\_GC , or run\_LP based on the user's specification on Task.

\begin{python}
def run_fedgraph(args, data):
    if args.fedgraph_task == "NC":
        run_NC(args, data)
    elif args.fedgraph_task == "GC":
        run_GC(args, data)
    elif args.fedgraph_task == "LP":
        run_LP(args)
\end{python}

\section{Supported Algorithms and Datasets}\label{sec:appendix-datasets}
For federated node classification, link prediction, and graph classification, we integrated different datasets and algorithms for each task, shown in Table~\ref{tab:supported_datasets} and Table~\ref{tab:supported_algorithm}. Researchers can also easily implement new algorithms and add their datasets.

\begin{table*}[h]
    \centering
\begin{tabular}{|c|c|}
\hline
\textbf{Task}        & \textbf{Dataset}                                                                                                                                    \\ \hline
Node Classification  & \begin{tabular}[c]{@{}c@{}}Cora, Citeseer, Pubmed, \\ Ogbn-Arxiv, Ogbn-Products, Ogbn-MAG~\cite{hu2020open}\end{tabular}                                              \\ \hline
Graph Classification & \begin{tabular}[c]{@{}c@{}}MUTAG, BZR, COX2, DHFR, \\ PTC-MR, AIDS, NCI1, ENZYMES, \\ DD, PROTEINS, COLLAB, \\ IMDB-BINARY, IMDB-MULTI~\cite{xie2021federated}\end{tabular} \\ \hline
Link Prediction      & FourSquare~\cite{yang2019revisiting}, WyzeRule~\cite{kamani2024wyze}                                                                                                                         \\ \hline
\end{tabular}
    \caption{Supported datasets of node classification, graph classification, and link prediction in federated learning.}
    \label{tab:supported_datasets}
\end{table*}

\begin{table*}[h]
    \centering
    \begin{tabular}{|c|c|c|}
\hline
\textbf{Task}                        & \textbf{Algorithm} & \textbf{Backbone} \\ \hline
\multirow{5}{*}{Node Classification} & FedAvg~\cite{li2019convergence}             & GCN               \\ \cline{2-3} 
                                     & Distributed GCN    & GCN               \\ \cline{2-3} 
                                     & BNS-GCN~\cite{wan2022bns}            & GCN               \\ \cline{2-3} 
                                     & FedSage+~\cite{zhang2021subgraph}           & GraphSage         \\ \cline{2-3} 
                                     & FedGCN~\cite{yao2024fedgcn}             & GCN,GraphSage     \\ \hline
\multirow{6}{*}{Graph Classification} & SelfTrain          & GIN               \\ \cline{2-3} 
                                     & FedAvg~\cite{li2019convergence}             & GIN               \\ \cline{2-3} 
                                     & FedProx~\cite{li2020federated}            & GIN               \\ \cline{2-3} 
                                     & GCFL~\cite{xie2021federated}               & GIN               \\ \cline{2-3} 
                                     & GCFL+~\cite{xie2021federated}              & GIN               \\ \cline{2-3} 
                                     & GCFL+dWs~\cite{xie2021federated}           & GIN               \\ \hline
\multirow{5}{*}{Link Prediction}     & FedAvg~\cite{li2019convergence}             & GCN               \\ \cline{2-3} 
                                     & STFL~\cite{lou2021stfl}               & GCN               \\ \cline{2-3} 
                                     & FedGNN+~\cite{gurler2022federated}            & GCN               \\ \cline{2-3} 
                                     & FedLink            & GCN               \\ \cline{2-3} 
                                     & FedRule~\cite{yao2023fedrule}            & GCN               \\ \hline
\end{tabular}
    \caption{Supported algorithms of node classification, graph classification, and link prediction in federated learning.}
    \label{tab:supported_algorithm}
\end{table*}



\section{Runner Workflow}\label{sec:appendix-workflow}

\subsection{Runner Workflow for Node Classification Task}
For the node classification task, the user specifies the dataset name in the configuration file. The dataset is preprocessed and partitioned across clients based on the chosen federated setting. The data is accessed directly from local storage or via an API, and then a time window may be generated to support temporal learning tasks.

Dataset extraction and preprocessing are managed by the function \texttt{dataloader\_NC} in the module \texttt{data\_process.py}, which handles both temporal and static graph datasets. Each client holds a subgraph or node features, and training proceeds in a federated manner.

The core execution is handled by \texttt{run\_NC}, which directs the process to an algorithm-specific runner \texttt{run\_NC\_\{algorithm\}}. Each algorithm may require different configurations, so separate \textit{.yaml} files are used. In each global round, clients perform local training, exchange model updates with the server, and participate in global validation. Results are recorded at the end of each round. The workflow is illustrated in Figure~\ref{fig:nc}.

\subsection{Runner Workflow for Graph Classification Task}

In the graph classification task, the dataset name can be specified by the user, and it could be either a single dataset or multiple datasets. For single dataset GC, the graphs will be assigned to a designated number of clients; for multiple datasets GC, each dataset will correspond to a client. Dataset generation and preparation are controlled by the function \texttt{dataloader\_GC} in the module \texttt{data\_process.py}. It will further assign the data process task to the function  \texttt{data\_loader\_GC\_single} or  \texttt{data\_loader\_GC\_multiple}. All the provided datasets are built-in TUDatasets in the Python library \texttt{torch\_geometric}.

For the graph classification task, different algorithms require different sets of arguments. Therefore, we divide the configurations into separate .yaml files, each corresponding to one algorithm. In \texttt{run\_GC}, the program will be further assigned to an algorithm-based runner \texttt{run\_GC\_\{algorithm\}}. The whole workflow is demonstrated in Figure \ref{fig:gc}.

\subsection{Runner Workflow for Link Prediction Task}

For the link prediction task, we provide a common dataset. The user only needs to specify the country codes. The original datasets are stored in Google Drive, so the user does not need to prepare the dataset by itself. The API will automatically check whether the dataset already exists and download the corresponding one if not.

\begin{figure*}[ht!]
    \centering
    \begin{subfigure}[b]{0.30\linewidth}
        \centering
        \includegraphics[width=\linewidth]{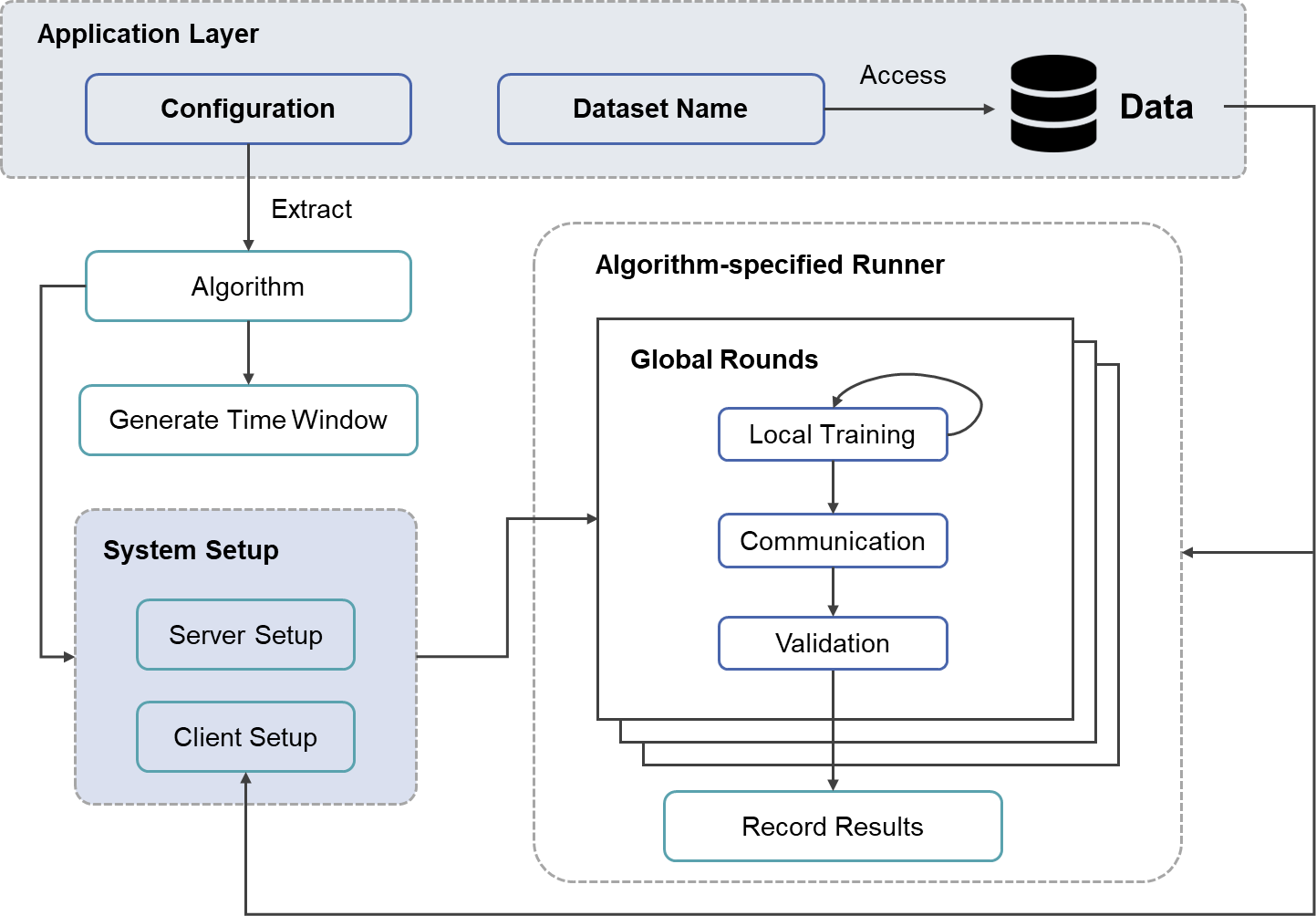}
        \caption{Node classification.}
        \label{fig:nc}
    \end{subfigure}
    \begin{subfigure}[b]{0.33\linewidth}
        \centering
        \includegraphics[width=\linewidth]{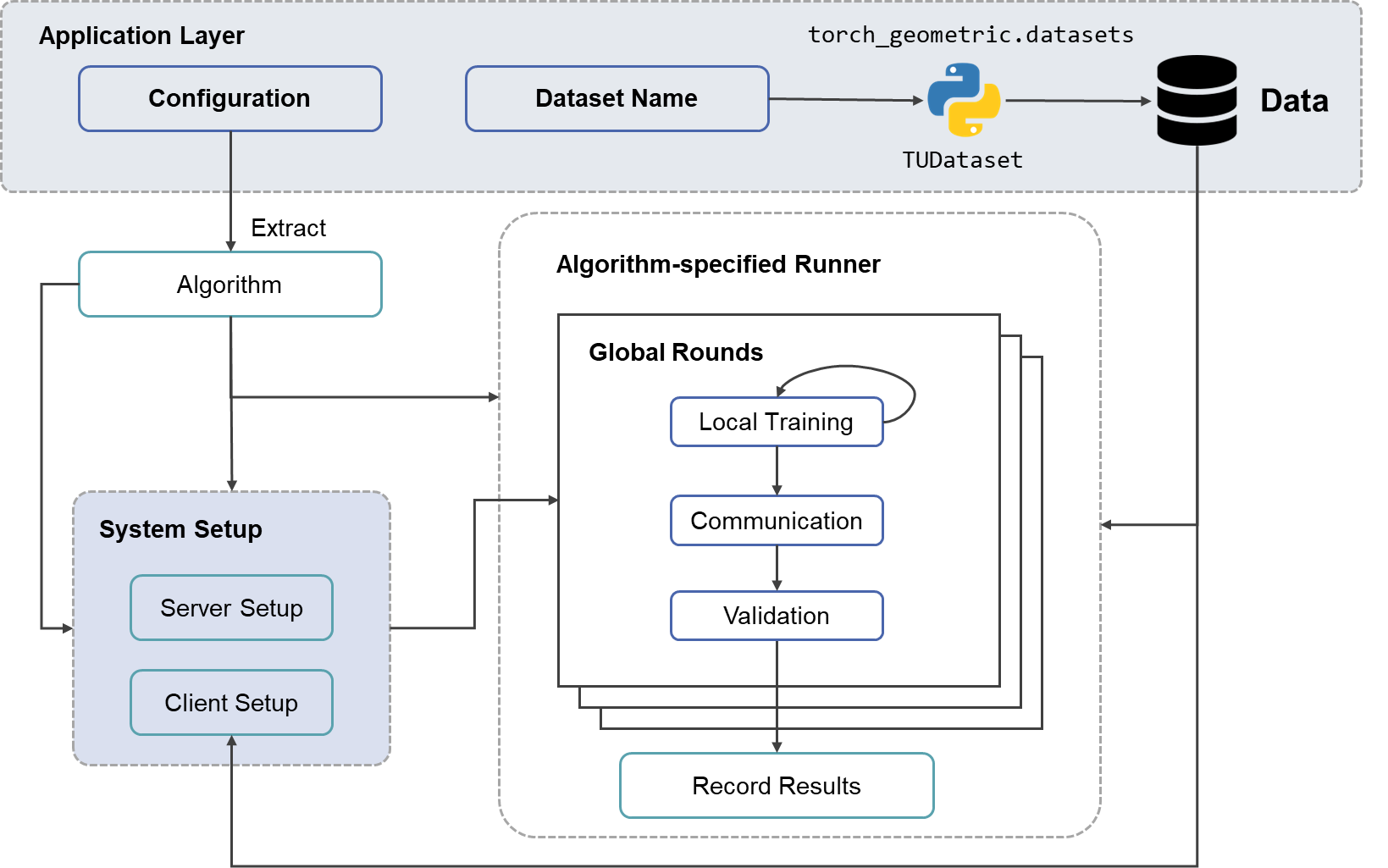}
        \caption{Graph classification.}
        \label{fig:gc}
    \end{subfigure}
    \begin{subfigure}[b]{0.29\linewidth}
        \centering
        \includegraphics[width=\linewidth]{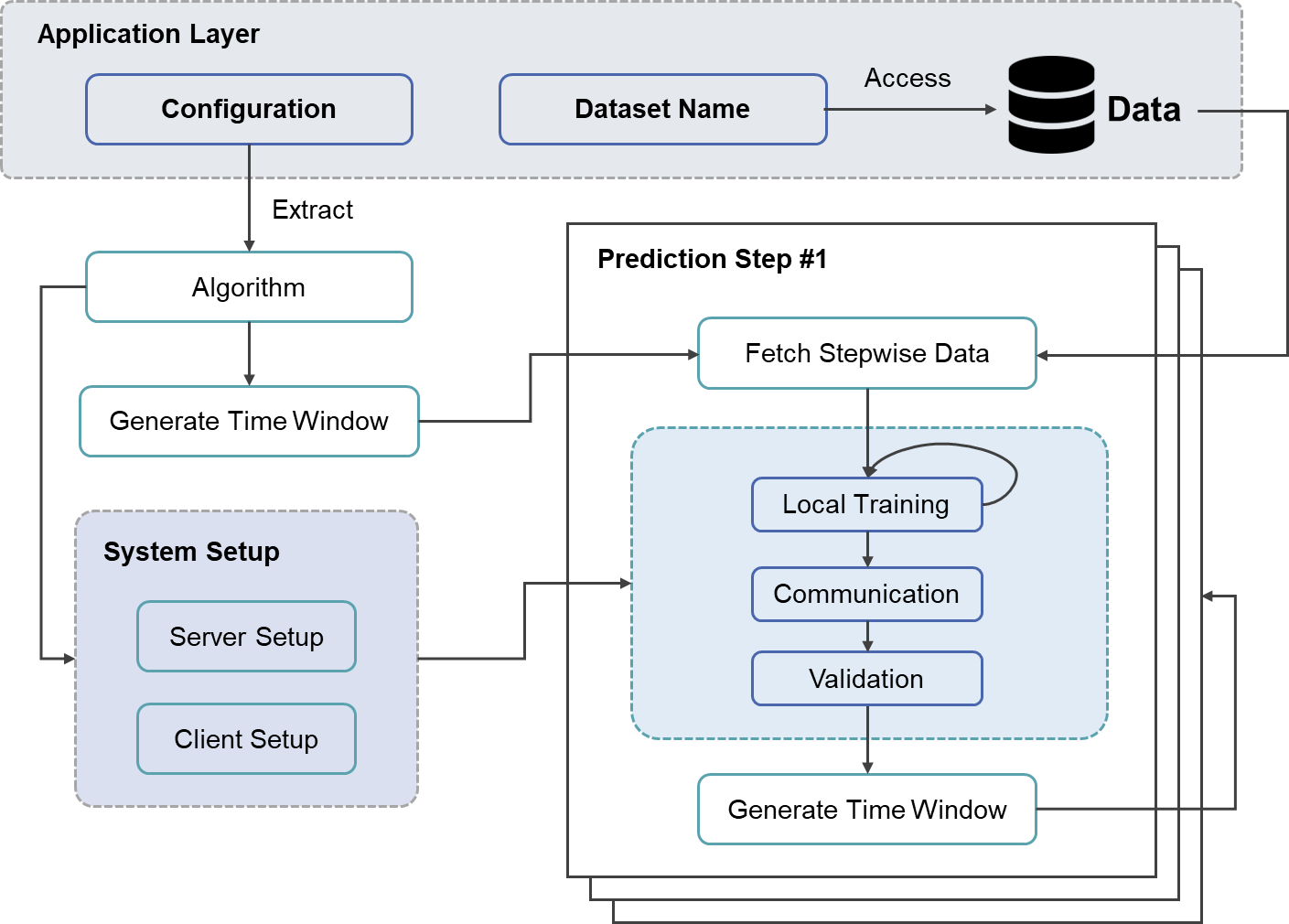}
        \caption{Link prediction.}
        \label{fig:lp}
    \end{subfigure}
    \caption{Federated graph training workflows for (a) node classification, (b) graph classification, and (c) link prediction tasks.}
    \label{fig:fedgraph_tasks}
\end{figure*}

There are also multiple available algorithms for the link prediction task. However, they share the same set of arguments so that we don't need to create separate \textit{.yaml} files. The user could conveniently select different algorithms by directly changing the algorithm field in the configuration file. All the algorithms will share the same runner. The whole workflow is demonstrated in Figure \ref{fig:lp}.

\section{Benchmark Homomorphic Encryption}\label{sec:appendix-he}

We provide a comprehensive guide for the configuration of Homomorphic Encryption in FedGraph to provide security guarantees corresponding to the graph structure and size. 
\subsection{Parameter Configuration}
In Table~\ref{tab:ckks_comparison}, we present the key parameters for configuring CKKS homomorphic encryption. The selection of the parameters is based on the dataset size and desired security level. Different parameter combinations create tradeoffs between computational overhead, communication cost, and precision. We present the default setting and parameter selection range to guide the user in selecting an appropriate combination that achieves the balance between objectives. 

\begin{table*}[ht]
\centering
\resizebox{1\textwidth}{!}
{%
\begin{tabular}{|p{2.8cm}|l|p{6.8cm}|p{2.8cm}|}
\hline
\textbf{Parameter} &
\textbf{Default Value} &
\textbf{Description} &
\textbf{Range} \\
\hline
scheme & CKKS & Encryption scheme type & N/A \\
\hline
polynomial modulus degree & 16384 & Maximum degree of polynomials used to represent encrypted data ($N \geq 2 \times $ max(nodes, features)) & 4096, 8192, 16384, 32768 \\
\hline
coefficient modulus bit size & [60, 40, 40, 40, 60] & Bit size for coefficient modulus that controls precision & Array of integers in the range $[20, 60]$ \\
\hline
global scale & $2^{40}$ & Global scale factor for encoding precision & $2^{30}$, $2^{40}$, $2^{50}$, etc. \\
\hline
security level & 128 & Bit security level & 128, 192, 256 \\
\hline
\end{tabular}%
}
\caption{TenSEAL Homomorphic Encryption Configuration Parameters. This table shows the key parameters for configuring the CKKS encryption scheme in the FedGraph library, including their default values, descriptions, and available ranges.}
\label{tab:ckks_comparison}
\end{table*}

\subsection{Microbenchmark}
We then provide the microbenchmark of HE on federated graph training in Table~\ref{tab:ckks_comparison_2}. The experiments are conducted on a 2-layer FedGCN for node classification tasks, running 100 global rounds with default settings in Cora. For CKKS parameters, we evaluate different polynomial modulus degrees (Poly\_mod), coefficient modulus sizes (Coeff\_mod), and precision levels. Time(s) for the encrypted version show pre-train/training/total times, respectively. Communication costs (Comm\_cost) include both pre-training and training rounds.

\textbf{Dynamic Precision:}  We adjust encryption parameters based on graph sizes and the numerical precision needed. For graphs like Cora, a polynomial modulus degree of 16384 with precision $2^{40}$ satisfies the ideal accuracy, while an increased value provides more precise security protection.    \\
\textbf{Communication Cost Optimization:} We employ several strategies to manage the communication overhead inherent in HE operation. The selection of coefficient modulus chain, [60,40,40,40,60], etc., enables efficient depth management for multiple HE operations. Depending on specific dataset characteristics (sparse matrix, larger datasets, etc.), we also employ efficient encryption methods to optimize communication cost and balance the performance.

When comparing among datasets, we observe that HE maintains equivalent accuracy across different parameter selections, as long as they satisfy the modulus requirement. If a smaller-than-required parameter size is used, the accuracy drops sharply, which indicates invalid encryption.





\begin{table*}[h]
\centering
\resizebox{1\textwidth}{!}
{%
\begin{tabular}{|p{1.5cm}|p{1cm}|c|c|c|c|p{1.5cm}|c|}
\hline
\textbf{Method} &  \textbf{Poly \_mod} & \textbf{Coeff\_mod} & \textbf{Precision} & \textbf{Dataset} & \textbf{Time(s)}& \textbf{Comm\_cost (MB)}& \textbf{Accuracy} \\
\hline
FedGCN (plaintext)  & N/A & N/A & N/A & Cora & 13.29 & 59.21 & 0.783 $\pm$ 0.07\\
\hline
FedGCN (HE) & 16384 & [60,40,40,40,60] & $2^{40}$ & Cora& 27.71/23.22/56.05 & 3279.15 & 0.779 $\pm$ 0.08 \\
\hline 
FedGCN (HE) & 32768 & [60,40,40,40,60] & $2^{50}$ & Cora & 29.44/36.17/71.76 & 4434.58 & 0.781 $\pm$ 0.08\\
\hline 
FedGCN (plaintext)  & N/A & N/A & N/A & Citeseer & 20.39 & 187.99 & 0.658 $\pm$ 0.06\\
\hline
FedGCN (HE) & 8192 & [60,40,40,60] & $2^{40}$ & Citeseer & 79.35/30.63/113.08 & 5791.42& 0.660 $\pm$ 0.07\\
\hline 
FedGCN (HE) & 16384 & [60,40,40,40,60] & $2^{40}$ & Citeseer & 123.22/45.8/173.40 & 8084.50& 0.652 $\pm$ 0.06\\
\hline
FedGCN (plaintext)  & N/A & N/A & N/A & PubMed & 25.78 & 150.43 & 0.774$\pm$ 0.12\\
\hline
FedGCN (HE) & 8192 & [60,40,40,60] & $2^{40}$ & PubMed & 70.28/19.73/93.18 & 3612.60& 0.757 $\pm$ 0.19\\
\hline
FedGCN (HE) & 16384 & [60,40,40,40,60] & $2^{40}$ & PubMed & 123.22/45.8/173.40 & 8084.50& 0.769 $\pm$ 0.13\\

\hline
\end{tabular}
}
\caption{Microbenchmark of FedGCN under Homomorphic Encryption with different CKKS scheme parameters. The experiments are conducted on a 2-layer FedGCN for node classification tasks, running 100 global rounds with default settings across three datasets (Cora, Citeseer, PubMed). For CKKS parameters, we evaluate different polynomial modulus degree (Poly\_mod), coefficient modulus sizes (Coeff\_mod), and precision levels. Time(s) for the encrypted version show pre-train/training/total times, respectively. Communication costs (Comm\_cost) include both pre-training and training rounds. Plain-text FedGCN serves as the baseline for comparison. Communication costs are measured for pre-training communication and training rounds separately. }  

\label{tab:ckks_comparison_2}
\end{table*}

\section{Additional Experiments}\label{sec:appendix-experiments}





\subsection{Increasing the Number of Clients with Fixed Computation Resources}
\begin{figure*}[ht]
    \centering
    \includegraphics[width=0.96\textwidth]{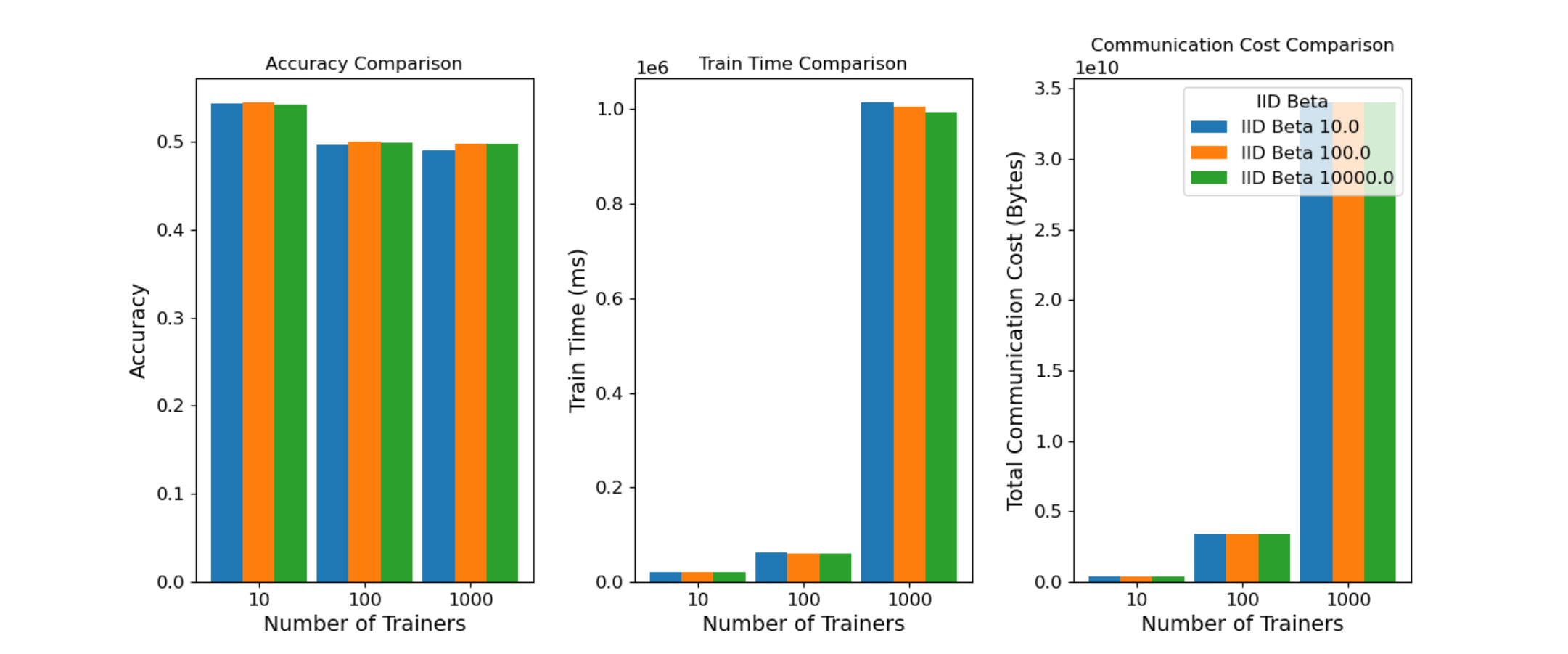}
    \caption{Training Time, Communication Cost, Test Accuracy on Ogbn-Arxiv in a Large Number of Clients. All experiments run on 10 AWS instances. 1000 trainers take a long time since it runs sequentially on 10 instances.}
    \label{fig:fedgraph_1000clients}
\end{figure*}
To better test scalability and fit real-world data, we increase the number of clients to 1000. In Figure \ref{fig:fedgraph_1000clients}, we observe that as the number of clients increases, the overall training time grows significantly due to sequential running on 10 instances, added communication overhead, and the need for increased synchronization among clients.

As we scale from 10 to 1000 clients under a fixed IID Beta value, there is a small decline in accuracy, likely due to the increased data heterogeneity each client possesses. The communication cost also escalates notably with more clients, highlighting the trade-off between parallelism and efficiency in federated settings. This experiment shows the system's ability to handle large-scale client distributions while revealing the resources required to maintain accuracy and efficiency.

\end{document}